%% file: main.tex
\pdfoutput=1

\documentclass[11pt]{article}

\usepackage{EMNLP2023}

\usepackage{times}
\usepackage{latexsym}

\usepackage[T1]{fontenc}

\usepackage[utf8]{inputenc}

\usepackage{microtype}

\usepackage{inconsolata}

\usepackage{amsmath}

\usepackage{times}
\usepackage{booktabs}
\usepackage{latexsym}
\usepackage{supertabular}
\usepackage{makecell}
\usepackage{multirow}
\usepackage{longtable}
\usepackage{subcaption}
\usepackage[inline]{enumitem}
\usepackage{graphicx}
\usepackage{array}
\usepackage{xspace}
\usepackage{afterpage}
\usepackage{balance}

\usepackage{soul}

\newcommand{\ctext}[3][RGB]{%
  \begingroup
  \definecolor{hlcolor}{#1}{#2}\sethlcolor{hlcolor}%
  \hl{#3}%
  \endgroup
}

\newif\ifhidecomments
\hidecommentsfalse

\ifhidecomments
  \newcommand{\chenhao}[1]{}
  \newcommand{\karen}[1]{}
\else
  \newcommand{\chenhao}[1]{\textcolor{blue}{[#1 ---\textsc{CT}]}}
  \newcommand{\karen}[1]{\textcolor{magenta}{[#1 ---\textsc{kz}]}}
  
\fi
\newcommand{\para}[1]{\noindent{\bf #1}}
\newcommand{\figref}[1]{Figure~\ref{#1}}
\newcommand{\tbref}[1]{Table~\ref{#1}}
\newcommand{\secref}[1]{\S\ref{#1}}

\newcommand{\PreSumm}{\textsc{\small PreSumm}\xspace}
\newcommand{\BART}{\textsc{\small BART}\xspace}
\newcommand{\PEGASUS}{\textsc{\small PEGASUS}\xspace}
\newcommand{\ProphetNet}{\textsc{\small ProphetNet}\xspace}

\newcommand{\TB}{Trump $\to$ Biden\xspace}
\newcommand{\BT}{Biden $\to$ Trump\xspace}

\title{
  Entity-Based Evaluation of Political Bias in Automatic Summarization
}

\author{Karen Zhou \\
  The University of Chicago\\
  \texttt{karenzhou@uchicago.edu} \\\And
  Chenhao Tan \\
  The University of Chicago\\
  \texttt{chenhao@uchicago.edu} \\}

\begin{document}
\maketitle
\begin{abstract}
  Growing literature has shown that NLP systems may encode social biases;
  however, the \textit{political} bias of summarization models remains relatively unknown. 
  In this work, we use an entity replacement method to investigate the portrayal of politicians in automatically generated summaries of news articles. 
  We develop an entity-based computational framework to assess the sensitivities of several extractive and abstractive summarizers to the politicians Donald Trump and Joe Biden.
  We find consistent differences in these summaries upon entity replacement, such as 
  reduced emphasis of Trump's presence in the context of the same article and
  a more individualistic representation of Trump with respect to the collective US government (i.e., administration). 
  These summary dissimilarities are most prominent when the entity is heavily featured in the source article.
  Our characterization provides a foundation for future studies of bias in summarization and for normative discussions on the ideal qualities of automatic summaries.
\end{abstract}

\input{intro.tex} %

\input{related.tex}

\input{data.tex}

\input{methods.tex}

\input{results.tex}

\input{conclusion.tex}

\section*{Acknowledgements}
We thank our anonymous reviewers for their helpful comments.
 We also thank Chenghao Yang, Shi Feng, and the other members of the Chicago Human+AI lab for their feedback and support.
 This work was in part supported by the AI \& democracy research initiative at the University of Chicago. Karen Zhou is additionally supported by a GFSD Fellowship.

\bibliography{references, refs}
\bibliographystyle{acl_natbib}

\appendix

\input{appendix_emnlp.tex}

\end{document}

%% file: intro.tex
\section{Introduction}

\begin{figure*}
	\centering
	\includegraphics[width=0.85\textwidth]{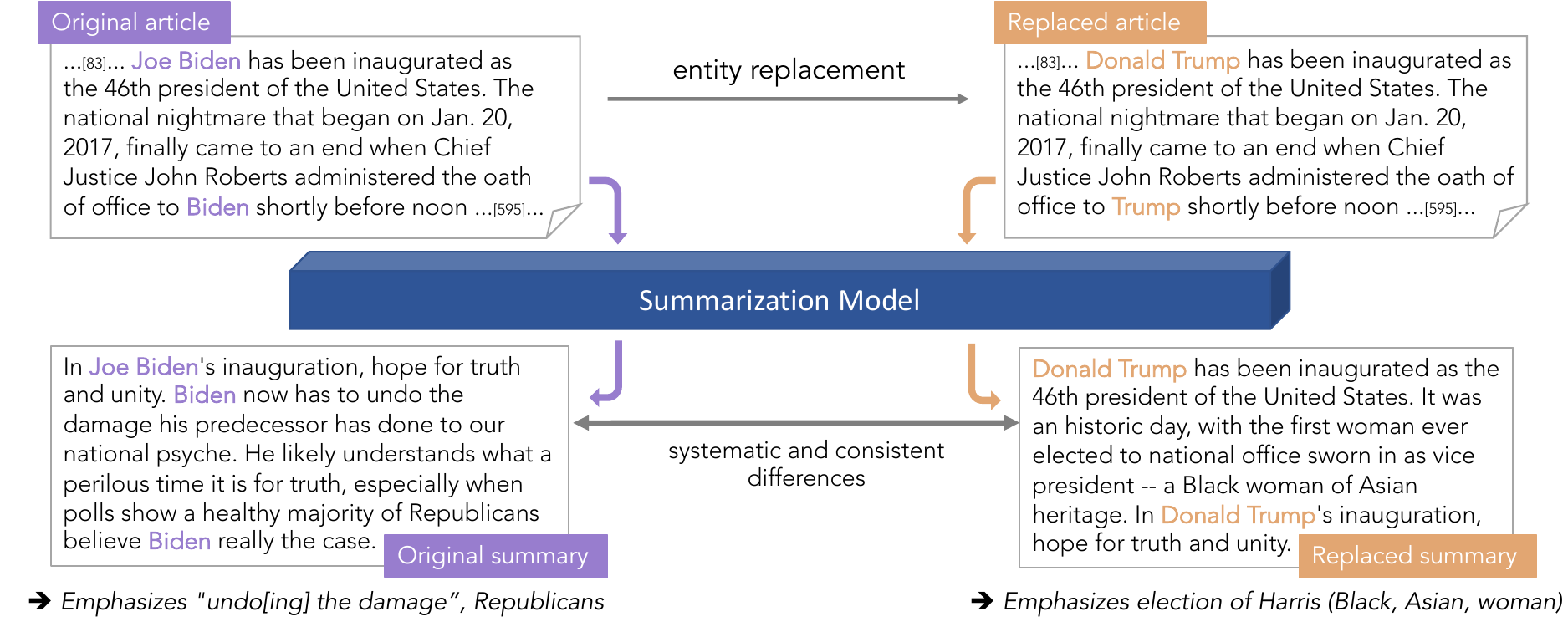}
	\caption{Illustration of our experimental design. Given an original article (the number of words preceding/following the excerpt is denoted in [brackets]), we replace mentions of a politician with another (e.g., Biden $\rightarrow$ Trump). This example shows that the model generates different summaries upon entity replacement in the source article. 
	}
	\label{fig:intro}
\end{figure*}

Automatic summarization aims to identify the {\em most important} information in input documents and produce a concise summary to save readers time and effort \citep{nenkova2011automatic}.
However, the pursuit of efficiency may come with undesired biases, especially in a political context. Bias of a political nature can contribute to issues like polarization and misinformation. 
It is also subjective and difficult to tell on the basis of a single article.

In this work, we conduct the first empirical study to examine the political bias of automatic summarizers based on large-scale pre-trained language models.
We investigate the portrayal of politicians in automatically generated summaries through an entity replacement experiment design; the advantage of this setup is the ability to control for the content of the source document.
\figref{fig:intro} shows an example: given a news article about Joe Biden, we replace all occurrences of Biden's name with that of Donald Trump and examine the differences between the summary of the original article and that of the replaced article.

What should an ideal summarization model output for these two versions of an article, {\em which differ only by one entity name}?
This normative question is crucial to consider for responsible deployment and usage of summarization models.
A first step towards addressing
this query is to uncover whether state-of-the-art automatic summarizers actually encode any bias at all.
In this paper, we focus on this empirical demonstration and leave the deliberation of desirable differences for future work (see \secref{sec:conc}).

We demonstrate substantial differences across four summarization models.
The consistent biases in the generated summaries
suggest robustly different model representations of the two politicians.
Furthermore, 
\emph{entity-centric} news articles, those that heavily feature the original entity, lead to more dissimilar summaries upon replacement.
{\bf We acknowledge that ``bias'' is an overloaded term;} in this paper, we use it to refer to {\bf significant differences with respect to entity representation.} {We do not endorse the interpretation that the summaries are ``anti-Trump/anti-conservative''.}

In summary, our main contributions are:
\begin{itemize}[itemsep=0pt]
    \item We conduct the first study on 
	the political
	bias
	in automatic summarization.
    \item We propose an entity replacement approach to compare, model representations of key political figures.
    \item We find consistent differences, such as under-emphasis of Trump's presence with the same article context and more individualistic representation of Trump in relation to the administration. These variations reveal that summarization models are not neutral with respect to political entities. 
\end{itemize}

%% file: related.tex
\section{Related Work}

There are two primary types of automatic summaries: \textit{extractive} 
and \textit{abstractive} 
\citep{el-kassas_automatic_2021}. 
Neural summarization models have demonstrated superior performance in benchmark datasets (e.g., \citet{zhang_pegasus_2020}).
To measure performance, ROUGE  scores are typically used \citep{lin_rouge_2004}, but such metrics have limited ability to cohesively assess summary quality \citep{maynez-etal-2020-faithfulness}, including any potential political bias.

There is strong resemblance between selecting the {\em most important} information in the summarization process and the phenomenon of media framing, where media outlets selectively report aspects of an issue or quotes of politicians \citep{entman1993framing,chong2007theory}.
The selection process by media outlets has received substantial attention in computational research \citep{niculae_quotus_2015,tan+peng+smith:18,guo_measuring_2022,vallejo2023connecting},
Our identified problem differs in that the content selection is done by models rather than media outlets and that the kind of bias may not align with the conservative vs. liberal biases as defined for media framing.

A growing line of work investigates the biases of NLP systems to understand their potential harms if/when deployed \citep{sheng-etal-2021-societal}.
Many studies have covered representation bias and social bias (e.g., gender and race) \citep{steen2023investigating, sheng_woman_2019, liang_towards_2021, yeo_defining_2020, garimella_he_2021, dhamala_bold_2021}.
In comparison, the harm of political bias is more subtle; it can be difficult to observe in an individual article, but can still have substantial impact
(e.g., affecting elections in US swing states). While \citet{feng-etal-2023-pretraining} tracks political bias in language models for hate speech and misinformation detection, our work is the first study to reveal political bias in automatic summarization models.

In addition, prior work has examined factual consistency in generated texts \citep{kryscinski_evaluating_2020, devaraj_evaluating_2022, Huang2021TheFI, maynez-etal-2020-faithfulness,tang_understanding_2022}, fairness of extractive summaries with classic summarization methods \citep{shandilya_fairness_2018}, and generating neutral summaries to mitigate media bias \citep{lee_neus_2022}. \citet{nan-etal-2021-entity} take an entity-based approach to evaluating factuality in abstractive summaries.

Replacement paradigms similar to our own have been used to uncover biases in other NLP applications, such as language modeling
\citep{feder_causalm_2021,vig_causal_2020, patel-pavlick-2021-stated} %
 and 
neural machine translation
\citep{wang_measuring_2022}.

%% file: data.tex
\section{Methodology}
\label{sec:data_and_methods}
In this section, we introduce our data,
experimental design, and the models we evaluate.
Data and code 
are available at \url{https://github.com/ChicagoHAI/entity-based-political-bias}.

\paragraph{Data:} For 
\textit{source documents},
we use cleanly-parsed
 articles from the NOW corpus\footnote{\url{https://www.english-corpora.org/now/}.} from 1/1/2020 to 12/31/2021 that have at most 5,000 words.
Among these news articles that mention any US politician, about 62.9\% include Donald Trump and/or Joe Biden.
Thus, we focus our investigation of bias on the entities Trump and Biden: well-known political figures that play important roles in US news. 
We separate the articles into two experimental groups:
157,648 articles that mention Trump only 
and 
85,952 articles that mention Biden only.\footnote{
During the $e_1 \to e_2$ entity replacement step, we consider only the articles that mention $e_1$ but not $e_2$.}
Appendix~\ref{sec:app_data_and_methods} contains additional statistics.

%% file: methods.tex
\begin{table}
    \small
    \centering 
    \begin{tabular}{c c }
        \toprule
        $p < 1e$-$20$:  & $\uparrow$$\uparrow$$\uparrow$$\uparrow$  $/$ $\downarrow$$\downarrow$$\downarrow$$\downarrow$  \\
        $p < 0.001$: & $\uparrow$$\uparrow$$\uparrow$  $/$ $\downarrow$$\downarrow$$\downarrow$  \\
        $p < 0.01$: & $\uparrow$$\uparrow$ $/$ $\downarrow$$\downarrow$ \\
        $p < 0.05$: & $\uparrow$ $/$ $\downarrow$\\
        $p \ge 0.05$: & \textemdash $/$ \textemdash \\
        \bottomrule
    \end{tabular}
    \caption{\label{arrows} Two-sided $t$-test thresholds and notations. The direction of the arrows indicates which class has the larger mean; i.e., $\uparrow$ indicates that $S^{e_2}_i$ has the larger mean, while $\downarrow$ corresponds to a larger mean in $S^{e_1}_i$.}
\end{table}

\paragraph{Replace and compare:}
Our experimental design consists of two steps.
Given a pair of politicians of interest ($e_1$, $e_2$),
we replace all the occurrences of $e_1$ in each news article with $e_2$
($e_1 \rightarrow e_2$ replacement).
We then have $N$ paired documents where the content only differs by mentions of these politicians: $\{(D_{i}^{e_1}, D_{i}^{e_2})_{i=1}^N\}$.
For a model $m$, we input each news article to $m$ and obtain summaries $\{(S_{i}^{e_1}, S_{i}^{e_2})_{i=1}^N\}_m$. 

We substitute each $e_1 \in \{\text{Trump, Biden}\}$ with each $e_2 \in \{\text{Trump, Biden, Obama, Bush}\} \setminus \{e_1\}$ in this paper. Presidents Barack Obama and George W. Bush are included as $e_2$ candidates to verify the robustness of our findings.
Appendix \ref{sec:app_data_and_methods} provides extra
details about these instantiations.

The second step is to assess the differences between these two sets of corresponding summaries $S_i^{e_1}$ and $S_i^{e_2}$. 
We define feature functions to extract properties of these summaries,
and then conduct paired $t$-tests to measure significance levels.
From the patterns arising in the results of these analyses, we develop an understanding of the models' encoded biases for representations of the entities.

\paragraph{Models:}
The summarization models used in this work include an extractive model 
and three abstractive models.
Specifically, we use the following:
\begin{itemize}[itemsep=0pt]
    \item \PreSumm \citep{liu_text_2019}: BERT-based model with hierarchical encoders, designed and fine-tuned for extractive text summarization.
    \item \PEGASUS \citep{zhang_pegasus_2020}: model with Gap Sentence Generation objective, pre-trained specifically for the abstractive summarization task. Training data $\sim$$46\times$ \PreSumm's.
	\item \BART \citep{lewis_bart_2019}: model with denoising corrupted input document objective, pre-trained for general NLG tasks. Training data $\sim$$10\times$ \PreSumm's.
    \item \ProphetNet \citep{qi_prophetnet_2020}: model with future n-gram prediction (using n-stream self-attention) objective, pre-trained for general sequence-to-sequence tasks. Training data $\sim$$10\times$ \PreSumm's.
\end{itemize}
All three abstractive models are Transformer-based encoder-decoders that are among the best performing with respect to ROUGE scores. 
They are also all fine-tuned for summarization on the CNN/DailyMail dataset \citep{hermann_teaching_2015}, consisting of news from 2007-2015. 
We use the default settings for all generation models as this is likely the most common use case in applications.

%% file: results.tex
\begin{figure}[t]
    \centering
    \includegraphics[width=0.45\textwidth]{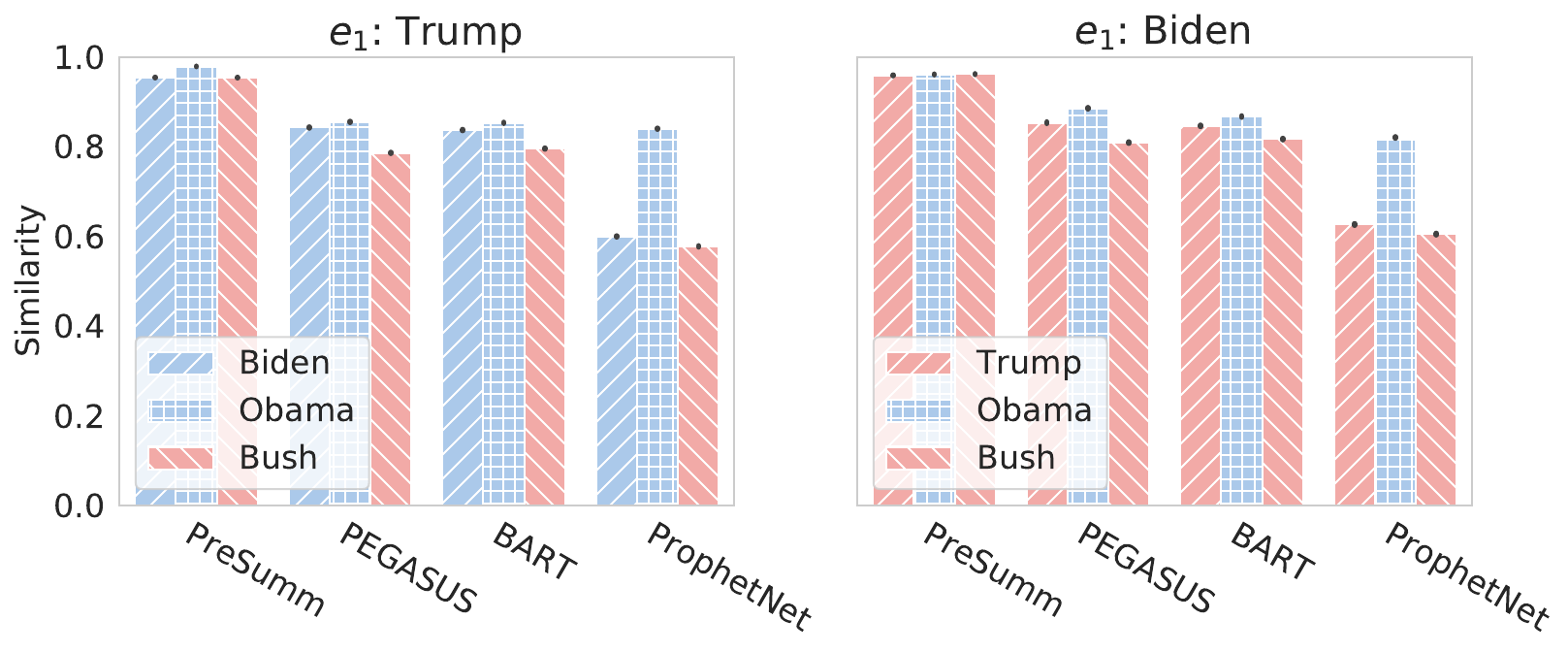}
    \caption{
        \PreSumm summaries have the highest similarity ratios, while \ProphetNet summaries are most dissimilar after entity replacement.
    }
    \label{fig:now_with_replace_similarity}
\end{figure}

\section{Framework and Results}
\label{sec:results}

We present our results for similarity between summaries, and two feature functions: 1) entity name mentions in summaries and 2) temporal differences in status indicators. 
We focus on these two features in this short paper to zero in on the basic notion of representation in sensitivity to political entities.

All results involve comparing frequencies via paired $t$-tests; we present results using the notation described in \tbref{arrows}. The numerical mean differences and $p$-values can be obtained from our publicly released code (see \secref{sec:data_and_methods}).
As described in Appendix \ref{ssec:lens}, there are small but signifcant differences in summary lengths before and after replacement. Therefore,
{all frequencies are normalized by length of summary} before the $t$-tests.

\subsection{Overall Similarity of $S_i^{e_1}$ and $S_i^{e_2}$}
We first compare the \textit{similarity scores} between the summary pairs.\footnote{We use \texttt{SequenceMatcher} from
\href{https://docs.python.org/3/library/difflib.html}{difflib}, which defines this score as the ratio $2.0*M / T$, where $T$ is the total number of words in both sequences, and $M$ is the number of matches, 
based on the Gestalt approach \cite{ratcliff1988pattern}.}
\figref{fig:now_with_replace_similarity} 
shows that \textsc{PreSumm} summaries have the highest similarity ratios as expected, since the summaries are directly extracted from the articles and the dissimilarity only stems from differences in selection.
\textsc{ProphetNet} summaries have the lowest similarities; this most powerful and latest summarization model before GPT-3 (see Limitations) is the most sensitive to differences in entity names, suggesting that it may encode the most differences between the two politicians.
Appendix \ref{sec:app_sim_comps} contains additional comparisons.

When investigating what factors are correlated with dissimilar summaries after entity replacement, we find that
{\bf entity-centric articles lead to more dissimilar summaries.}
From the \TB and \BT scores, we split the news articles into ones leading to highly similar summaries 
(\textsc{HighSim}) vs. those with highly dissimilar summaries (\textsc{LowSim}) for each model.\footnote{
We define these thresholds using the 75th- and 25th-percentiles 
respectively.} 
After creating each model's \textsc{HighSim} and \textsc{LowSim} article subsets, we identified distinguishing key words with the Fightin’ Words algorithm with the uninformative Dirichlet prior \citep{monroe_fightin_2009}.
We observe that the greater presence of $e_1$'s name consistently leads to dissimilar summaries for the $e_1 \to e_2$ directions. This indicates that \emph{entity-centric} articles lead to dissimilar summaries upon replacement; i.e.,
the model's representational differences emerge as Trump or Biden becomes the focus of the article. 
See Appendix \ref{sec:app_sim_comps} for detailed results from this exploration.

\begin{table}[t]
    \centering
    \small
    \begin{tabular}{p{1.5cm} p{0.5cm} p{0.6cm}p{0.5cm} p{0.5cm} p{0.6cm} p{0.5cm}}
        \toprule
        {$e_1$} &\multicolumn{3}{c}{Trump} &         \multicolumn{3}{c}{Biden}   \\
        \cmidrule(lr){2-4} \cmidrule(lr){5-7}
        {$e_2$} &    Biden &                                   Obama &                          Bush &                                    Trump &                                   Obama &                                   Bush \\
        \midrule
        \PreSumm             &            $\uparrow$$\uparrow$$\uparrow$ &  $\downarrow$$\downarrow$$\downarrow$$\downarrow$ &                      $\uparrow$$\uparrow$ &                                       \textemdash &  $\downarrow$$\downarrow$$\downarrow$$\downarrow$ &              $\downarrow$$\downarrow$$\downarrow$ \\
\PEGASUS &  $\uparrow$$\uparrow$$\uparrow$$\uparrow$ &          $\uparrow$$\uparrow$$\uparrow$$\uparrow$ &  $\uparrow$$\uparrow$$\uparrow$$\uparrow$ &  $\downarrow$$\downarrow$$\downarrow$$\downarrow$ &  $\downarrow$$\downarrow$$\downarrow$$\downarrow$ &  $\downarrow$$\downarrow$$\downarrow$$\downarrow$ \\
\BART        &  $\uparrow$$\uparrow$$\uparrow$$\uparrow$ &          $\uparrow$$\uparrow$$\uparrow$$\uparrow$ &  $\uparrow$$\uparrow$$\uparrow$$\uparrow$ &  $\downarrow$$\downarrow$$\downarrow$$\downarrow$ &              $\downarrow$$\downarrow$$\downarrow$ &          $\uparrow$$\uparrow$$\uparrow$$\uparrow$ \\
\ProphetNet  &  $\uparrow$$\uparrow$$\uparrow$$\uparrow$ &          $\uparrow$$\uparrow$$\uparrow$$\uparrow$ &  $\uparrow$$\uparrow$$\uparrow$$\uparrow$ &  $\downarrow$$\downarrow$$\downarrow$$\downarrow$ &          $\uparrow$$\uparrow$$\uparrow$$\uparrow$ &          $\uparrow$$\uparrow$$\uparrow$$\uparrow$ \\
        \bottomrule
        \end{tabular}
        \caption{\label{tab:uncond_freqs_ttest}
        Trump is less likely to be mentioned in abstractive summaries than the other presidents, suggesting that these models exhibit inclusion bias. The consistent $\uparrow$ arrows for Trump $\to$ $e_2$ show that replacing Trump's name with $e_2$ in the source article results in abstractive summaries with significantly higher frequencies of $e_2$; this observation appears consistent for Biden $\to$ Trump.
        }
\end{table}

\input{freqs.tex}

\input{transition.tex}

%% file: freqs.tex
\subsection{Entity Representation in Summaries} 

We then determine whether the summary pairs mention their entities to the same extent. 
This property reflects \textit{inclusion bias}~\citep{steen2023investigating}: the idea that entities should be mentioned in a summary with equal frequency if they are equally salient in the source document. 
We define this feature function as the frequency with which the entity is named in the summary, 
i.e., $\#e_1$ in summary, 
and likewise for $e_2$ after replacement.\footnote{To ensure fair comparison, we restore $e_1$ in $S_i^{e_2}$ where it makes sense.}

\paragraph{Trump is least likely to be mentioned in abstractive summaries.} 
The abstractive summarizers tend \emph{not} to mention Trump's name
(see \tbref{tab:uncond_freqs_ttest}).  
There are also significant differences in usage of Biden's name compared to the other presidents.
This result suggests that such summaries are sensistive to the entities in the article. In particular, they de-emphasize Trump's presence compared to other presidents in the context of the same article, which may skew reader perception of the event. See Appendix \ref{app:ex} for example summary pairs.

%% file: transition.tex
\begin{table*}
    \small
    \centering
    \begin{tabular}{lllllllllllll}
        \toprule
        {$e_1$} & \multicolumn{6}{c}{Trump} & \multicolumn{6}{c}{Biden}  \\
        \cmidrule(lr){2-7} \cmidrule(lr){8-13}
        {$e_2$} & \multicolumn{2}{c}{Biden} & \multicolumn{2}{c}{Obama} & \multicolumn{2}{c}{Bush} & \multicolumn{2}{c}{ Trump} & \multicolumn{2}{c}{Obama} & \multicolumn{2}{c}{Bush} \\
        {} &                                        20 &                                        21 &                    20 &           21 &                              20 &           21 &                                    20 &                                    21 &                                    20 &                                    21 &                                    20 &                                                21 \\
        \midrule
        \PreSumm             &                                $\uparrow$ &                      $\uparrow$$\uparrow$ &                     \textemdash &  \textemdash &                     \textemdash &  \textemdash &  $\downarrow$$\downarrow$$\downarrow$ &                           \textemdash &  $\downarrow$$\downarrow$$\downarrow$ &  $\downarrow$$\downarrow$$\downarrow$ &              $\downarrow$ &                          $\downarrow$ \\
\PEGASUS &  $\uparrow$$\uparrow$$\uparrow$$\uparrow$ &  $\uparrow$$\uparrow$$\uparrow$$\uparrow$ &                     \textemdash &  \textemdash &                    $\downarrow$ &  \textemdash &  $\downarrow$$\downarrow$$\downarrow$ &  $\downarrow$$\downarrow$$\downarrow$ &  $\downarrow$$\downarrow$$\downarrow$ &  $\downarrow$$\downarrow$$\downarrow$ &              $\downarrow$ &  $\downarrow$$\downarrow$$\downarrow$ \\
\BART       &            $\uparrow$$\uparrow$$\uparrow$ &            $\uparrow$$\uparrow$$\uparrow$ &  $\uparrow$$\uparrow$$\uparrow$ &  \textemdash &  $\uparrow$$\uparrow$$\uparrow$ &  \textemdash &              $\downarrow$$\downarrow$ &              $\downarrow$$\downarrow$ &  $\downarrow$$\downarrow$$\downarrow$ &              $\downarrow$$\downarrow$ &                $\uparrow$ &                           \textemdash \\
\ProphetNet  &            $\uparrow$$\uparrow$$\uparrow$ &            $\uparrow$$\uparrow$$\uparrow$ &                     \textemdash &  \textemdash &                     \textemdash &  \textemdash &  $\downarrow$$\downarrow$$\downarrow$ &                          $\downarrow$ &  $\downarrow$$\downarrow$$\downarrow$ &  $\downarrow$$\downarrow$$\downarrow$ &  $\downarrow$$\downarrow$ &  $\downarrow$$\downarrow$$\downarrow$ \\

        \bottomrule
        \end{tabular}
        \caption{\label{tab:norm_vice_year} 
		Biden is more likely to be associated with ``Vice President'' in nearly all cases. 
        The consistent $\downarrow$ arrows for Biden $\to e_2$ show that replacing Biden's name with $e_2$ in the source article results in summaries with significantly lower inclusion frequencies of ``Vice President''; the phenomenon is consistent for Trump $\to$ Biden.
		}
\end{table*}

\begin{table*}
    \centering
    \small
    \begin{tabular}{lllllllllllll}
        \toprule
        {$e_1$} & \multicolumn{6}{c}{Trump} & \multicolumn{6}{c}{Biden}  \\
        \cmidrule(lr){2-7} \cmidrule(lr){8-13}
        {$e_2$} & \multicolumn{2}{c}{Biden} & \multicolumn{2}{c}{Obama} & \multicolumn{2}{c}{Bush} & \multicolumn{2}{c}{ Trump} & \multicolumn{2}{c}{Obama} & \multicolumn{2}{c}{Bush} \\
        {} &                                        20 &                                    21 &                                        20 &                              21 &                                        20 &                              21 &           20 &                                                21 &           20 &                              21 &                                    20 &                                                21 \\
        \midrule
        \PreSumm             &      $\downarrow$$\downarrow$$\downarrow$ &  $\downarrow$$\downarrow$$\downarrow$ &                               \textemdash &                    $\downarrow$ &                                $\uparrow$ &                     \textemdash &  \textemdash &          $\uparrow$$\uparrow$$\uparrow$$\uparrow$ &                           \textemdash &                    $\uparrow$$\uparrow$$\uparrow$ &                           \textemdash &          $\uparrow$$\uparrow$$\uparrow$$\uparrow$ \\
        \PEGASUS &  $\uparrow$$\uparrow$$\uparrow$$\uparrow$ &        $\uparrow$$\uparrow$$\uparrow$ &  $\uparrow$$\uparrow$$\uparrow$$\uparrow$ &        $\downarrow$$\downarrow$ &  $\uparrow$$\uparrow$$\uparrow$$\uparrow$ &                     \textemdash &  \textemdash &  $\downarrow$$\downarrow$$\downarrow$$\downarrow$ &  $\downarrow$$\downarrow$$\downarrow$ &  $\downarrow$$\downarrow$$\downarrow$$\downarrow$ &  $\downarrow$$\downarrow$$\downarrow$ &              $\downarrow$$\downarrow$$\downarrow$ \\
        \BART        &  $\uparrow$$\uparrow$$\uparrow$$\uparrow$ &                  $\uparrow$$\uparrow$ &  $\uparrow$$\uparrow$$\uparrow$$\uparrow$ &  $\uparrow$$\uparrow$$\uparrow$ &  $\uparrow$$\uparrow$$\uparrow$$\uparrow$ &  $\uparrow$$\uparrow$$\uparrow$ &  \textemdash &  $\downarrow$$\downarrow$$\downarrow$$\downarrow$ &                           \textemdash &                                      $\downarrow$ &                           \textemdash &                    $\uparrow$$\uparrow$$\uparrow$ \\
        \ProphetNet  &  $\uparrow$$\uparrow$$\uparrow$$\uparrow$ &        $\uparrow$$\uparrow$$\uparrow$ &  $\uparrow$$\uparrow$$\uparrow$$\uparrow$ &  $\uparrow$$\uparrow$$\uparrow$ &  $\uparrow$$\uparrow$$\uparrow$$\uparrow$ &                     \textemdash &  \textemdash &  $\downarrow$$\downarrow$$\downarrow$$\downarrow$ &              $\downarrow$$\downarrow$ &                                       \textemdash &                           \textemdash &  $\downarrow$$\downarrow$$\downarrow$$\downarrow$ \\        
        \bottomrule
        \end{tabular}
        \caption{\label{tab:norm_admin} 
		Trump is less likely to appear with the word ``administration'' in all abstractive summaries, suggesting that those models hold a less unified view of him
        with respect to the US government than the other presidents. The consistent $\uparrow$  arrows for Trump $\to e_2$ show that replacing Trump's name with $e_2$ in the source article results in summaries with significantly more mentions of ``administration''; this observation is consistent for Biden $\to$ Trump.
		}
\end{table*}

\subsection{Indicators of Status Over Time
}

An important event during the time period of our data is the inauguration of Biden at the start of 2021, which changes his status from former Vice President to President.
To identify changes across time due to this event,
we separate the articles by year of publication (2020 vs.\ 2021), and then assess the frequency of the title ``Vice President'' across generated summaries $S_i^{e_1}$ and $S_i^{e_2}$.

Another relevant change is the use of the word ``administration'' (e.g., ``the Biden administration''), which can be seen as an indicator of whether the presidency is viewed as an individualistic or a collective effort. 
\citet{sunstein_president_1994} also draw distinctions between the president and the administration, and their perceived authority. 
The frequency of this term is another feature function.

\paragraph{Most models associate Biden with being the Vice President.}
From the frequency of ``Vice President'' usage in 2020 vs.\ 2021 (\tbref{tab:norm_vice_year}), Biden is more strongly associated with the title of ``Vice President''. This observation holds even in 2021 summaries, when he assumes 
the title ``President''.
Our observation is aligned with recent work on hallucination and factuality \citep{ji2022survey,maynez_faithfulness_2020},
highlighting their potential impacts in 
political news summarization.

\paragraph{Abstractive models tend not to associate Trump with the ``administration''.}
The abstractive models consistently dissociate Trump with the use of ``administration''  (Table~\ref{tab:norm_admin}). This result suggests that Trump's actions may not be representative of those of the broader US government; in other words,
 these models encode a more individualistic representation of Trump compared to other presidents. 
Examples of these results are in Appendix \ref{app:ex}.

%% file: conclusion.tex
\section{Conclusion} 
\label{sec:conc}

In this work, we conduct the first empirical study on political bias in summarization models.
Through experiments that replace the names of Donald Trump and Joe Biden with other presidential entities in US news articles, we identify several areas of variation in summaries when the source article differs by a single person's name.
We show that {\bf summarization models learn different representations for different entities and provide consistently different summaries depending on the entities involved}.
Our framework can be used for further analysis of 
biases in summarization models.

\input{discussion.tex}

\section*{Limitations} 
We acknowledge the limitations of our methodology, which provide opportunities for further research.
This proposed framework was only evaluated on US politicians, specifically two US presidents. 
There are limits to studying only these two entities, and not other less popular/controversial politicians. We attempt to account for this by adding the additional $e_2$ candidates Obama and Bush.
Additionally, because Trump and Biden are prominent political figures that feature heavily in US political news (see \secref{sec:data_and_methods}), the consistent biases we found across multiple summarization models 
constitute important initial findings. Summaries of such high-profile politicians are more likely to be read and have possible harms. 
Furthermore, different from the emphasis on generalization in modeling work, scholars in other fields such as political science may dedicate an entire book to the study of a single politician (e.g., \citet{wilson2016barack}).

Political bias may vary across different countries and regimes.
We also do not claim that the observed differences are exclusive to the two politicians that we study here.
They likely exist for any notable entities in the models' training data.

Another limitation lies in our work's exclusive focus on English news articles.
While the frequency metrics can generalize to other nations and languages, some analyses done here may be specific to English terms and dynamics of US politics. 
We also do not consider co-reference resolution of politician titles in entity replacement.
Entity-swapping between politicians of different genders may require additional data processing due to the need for pronoun resolution.

Our work uses the default generation setup in the four models of interest.
We recognize that there exist a multitude of hyperparameter settings and decoding algorithms that can lead to summaries of different properties.
Future studies can also examine the impact of decoding algorithms on biases in automatic summarization.

Finally, while GPT-3+ has been shown to perform well on some summarization tasks \citep{Goyal2022NewsSA, Yang2023ExploringTL, Zhang2023ExtractiveSV}, we do not include it since our study began prior to the API's accessibility. Furthermore, the models we evaluated are publicly available, which allows for greater control over the generation of summaries. As political bias is a sensitive social issue with subjective connotations, we believe that it is important to conduct a fair and thorough comparison in a controllable setting.
Our proposed framework can be readily applied to GPT-3+ and other models in future evaluations.

\section*{Ethics Statement}
Having demonstrated systematic biases related to entity replacement, one could knowingly manipulate a summary of a news article about Trump by replacing his name with Biden's, and then restoring his original name in the adversarial summary. This action is related to spinned summaries~\citep{bagdasaryan_spinning_2021}, and there may be concerns about contributing to fake news and disinformation \citep{buchanan_truth_2021}.  %

On the other hand, this research contributes to revealing potential biases and harms of NLP systems.
A primary benefit of our findings is improving public understanding of NLP models, which may lead to cautious deployment of 
such models.
We believe that this is especially important for summarization, which is often portrayed as a straightforward task.
Ultimately, we believe that the benefits of being able to characterize political biases in automatic summarization will aid in developing defenses to such adversarial use cases.
Raising awareness of such biases can also enable an informed discussion on what desirable summaries should include and how summarization models should be used.

Overall, our work adds to the emerging area of understanding potential issues of NLP systems, and it makes an important contribution towards recognizing the political biases of summarization models, which is critical for their responsible use.

%% file: discussion.tex
Politicians are often in similar situations, such as signing a bill or meeting a world leader. Models with distinct representations for different entities will generate different summaries for such reports. 
These variations challenge the assumption that summaries are inherently neutral, that
one can produce or seek a
single gold-standard automatic summary.
Such differences and their downstream effects (e.g., on voter preferences)
warrant further exploration.

Ultimately, we must consider: {\bf what constitutes an ideal automatic summary?}
This is an important normative query, which we purposefully do not answer in this paper. 
Having distinct representations of politicians may indicate a nuanced model, but that does not automatically justify
all differences.
Characterizing the systematic biases encoded in automatic summarizers is a first step towards responsible use of such models.
Our work aims to motivate discussion of this essential yet overlooked research area.

%% file: appendix_emnlp.tex
\appendix

\section{Methodology Details}
\label{sec:app_data_and_methods}
\begin{figure}[t]
	\centering

	\includegraphics[width=0.4\textwidth]{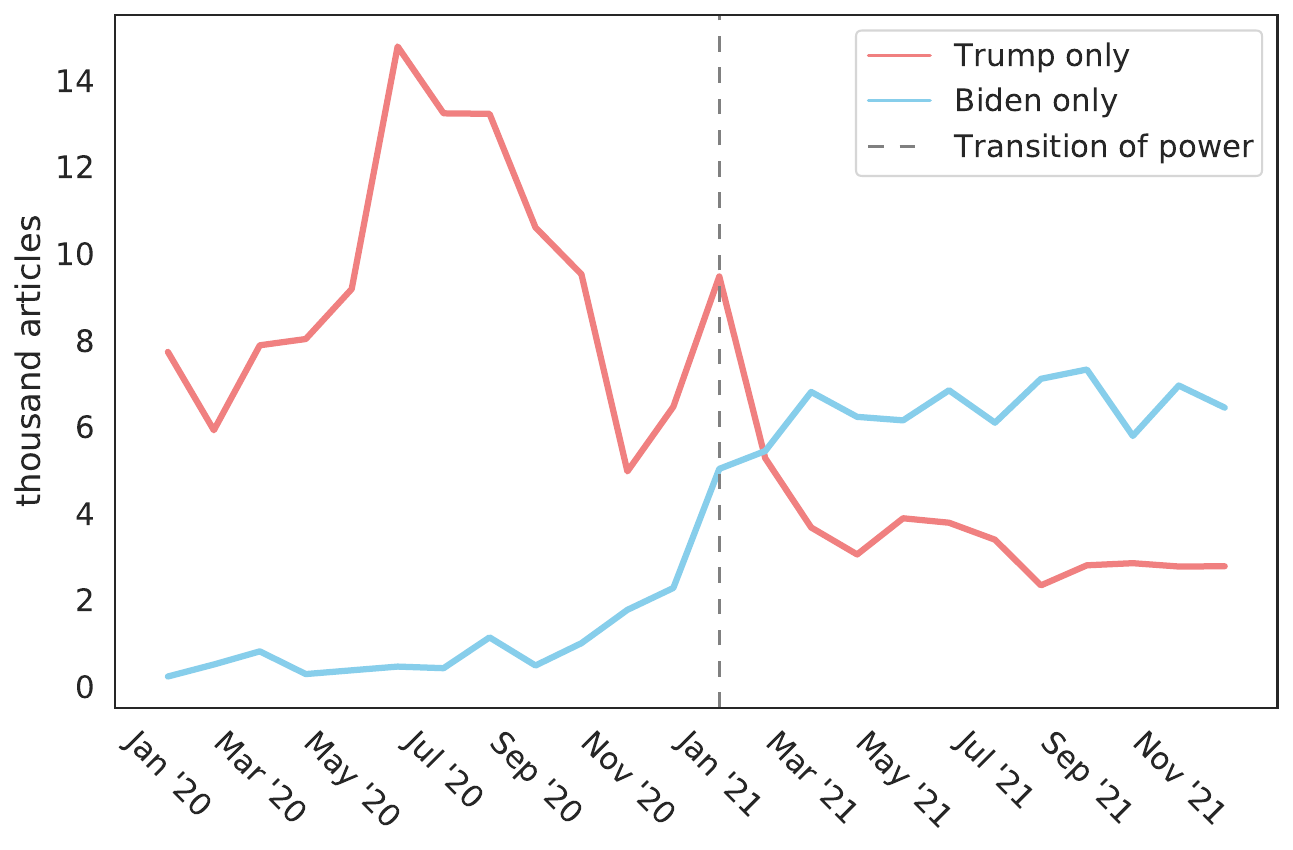}

	\caption{Number of articles by month. The quantity of ``Trump only'' articles declines after the inauguration in January 2021, while ``Biden only'' articles increase. 
	}
	\label{fig:data_hist}
\end{figure}

\para{Data statistics:} The ``Trump only'' articles average 744.8 words in length and the ``Biden only'' articles average 808.0 words in length. \figref{fig:data_hist} shows the distribution of articles that include these entity mentions. ``Trump only'' articles decrease in number after the inauguration in January 2021, while the number of ``Biden only'' articles increases. 

As stated in \secref{sec:data_and_methods}, among 2020-2021 news articles that mention any US politician, about 62.9\% include Donald Trump and/or Joe Biden. We filter the articles mentioning any US politician as ones that include names from \href{https://github.com/CivilServiceUSA/}{CivilServiceUSA}. Then, among those articles, we search for the fraction that also mention ``Trump'' and ``Biden''.

\begin{table}[t]
    \small
    \centering 
    \setlength\extrarowheight{2pt}
    \begin{tabular}{ >{\centering\arraybackslash}p{2.5cm} c  >{\centering\arraybackslash}p{3.5cm} }
        \hline
         Trump & $\leftrightarrow$ & Biden \\
        \hline
        {Donald Trump, D. Trump} & 	$\leftrightarrow$  & {Joe Biden, Joseph Biden, J. Biden} \\
        \hline
        {Donald J. Trump, Donald John Trump} &	$\leftrightarrow$  & {Joe R. Biden, Joseph R. Biden, Joe Robinette Biden, Joseph Robinette Biden}\\
        \hline
    \end{tabular}
    \caption{Mapping between names for the entity replacement. If the name form has multiple mappings (e.g., ``Donald Trump'' $\to$ \{``Joe Biden'', ``Joseph Biden''\}), the first name in the list is used as the default value. 
	}
	\label{tab:replace}
\end{table}

\begin{table}
	\centering
	\small
	\begin{tabular}{p{1.5cm} p{0.5cm} p{0.6cm}p{0.5cm} p{0.5cm} p{0.6cm} p{0.5cm}}
		\toprule
        {$e_1$} &\multicolumn{3}{c}{Trump} &         \multicolumn{3}{c}{Biden}   \\
        \cmidrule(lr){2-4} \cmidrule(lr){5-7}
        {$e_2$} &    Biden &                                   Obama &                          Bush &                                    Trump &                                   Obama &                                   Bush \\
        \midrule
		\PreSumm             &                    $\uparrow$$\uparrow$$\uparrow$ &              $\downarrow$$\downarrow$$\downarrow$ &          $\uparrow$$\uparrow$$\uparrow$$\uparrow$ &      $\downarrow$$\downarrow$$\downarrow$ &      $\downarrow$$\downarrow$$\downarrow$ &                    $\uparrow$$\uparrow$$\uparrow$ \\
\PEGASUS &  $\downarrow$$\downarrow$$\downarrow$$\downarrow$ &  $\downarrow$$\downarrow$$\downarrow$$\downarrow$ &  $\downarrow$$\downarrow$$\downarrow$$\downarrow$ &  $\uparrow$$\uparrow$$\uparrow$$\uparrow$ &                  $\downarrow$$\downarrow$ &              $\downarrow$$\downarrow$$\downarrow$ \\
\BART       &              $\downarrow$$\downarrow$$\downarrow$ &              $\downarrow$$\downarrow$$\downarrow$ &  $\downarrow$$\downarrow$$\downarrow$$\downarrow$ &            $\uparrow$$\uparrow$$\uparrow$ &                               \textemdash &  $\downarrow$$\downarrow$$\downarrow$$\downarrow$ \\
\ProphetNet &  $\downarrow$$\downarrow$$\downarrow$$\downarrow$ &          $\uparrow$$\uparrow$$\uparrow$$\uparrow$ &          $\uparrow$$\uparrow$$\uparrow$$\uparrow$ &                               \textemdash &  $\uparrow$$\uparrow$$\uparrow$$\uparrow$ &          $\uparrow$$\uparrow$$\uparrow$$\uparrow$ \\

		\bottomrule
		\end{tabular}
		\caption{Summary length $t$-test results. Abstractive summaries tend to be longer for Trump. 
	}
	\label{tab:app_sum_ttest}
		
\end{table}

\para{Additional replace and compare details:}
To do the $e_1 \to e_2$ replacement, we create mappings between the different variations of the entity names.
In the case of Trump and Biden, we use the name mapping in Table~\ref{tab:replace} to do the \TB  and \BT replacements. Similar mappings are utilized for Obama and Bush.

\begin{figure}
	\centering

		\includegraphics[width=0.35\textwidth]{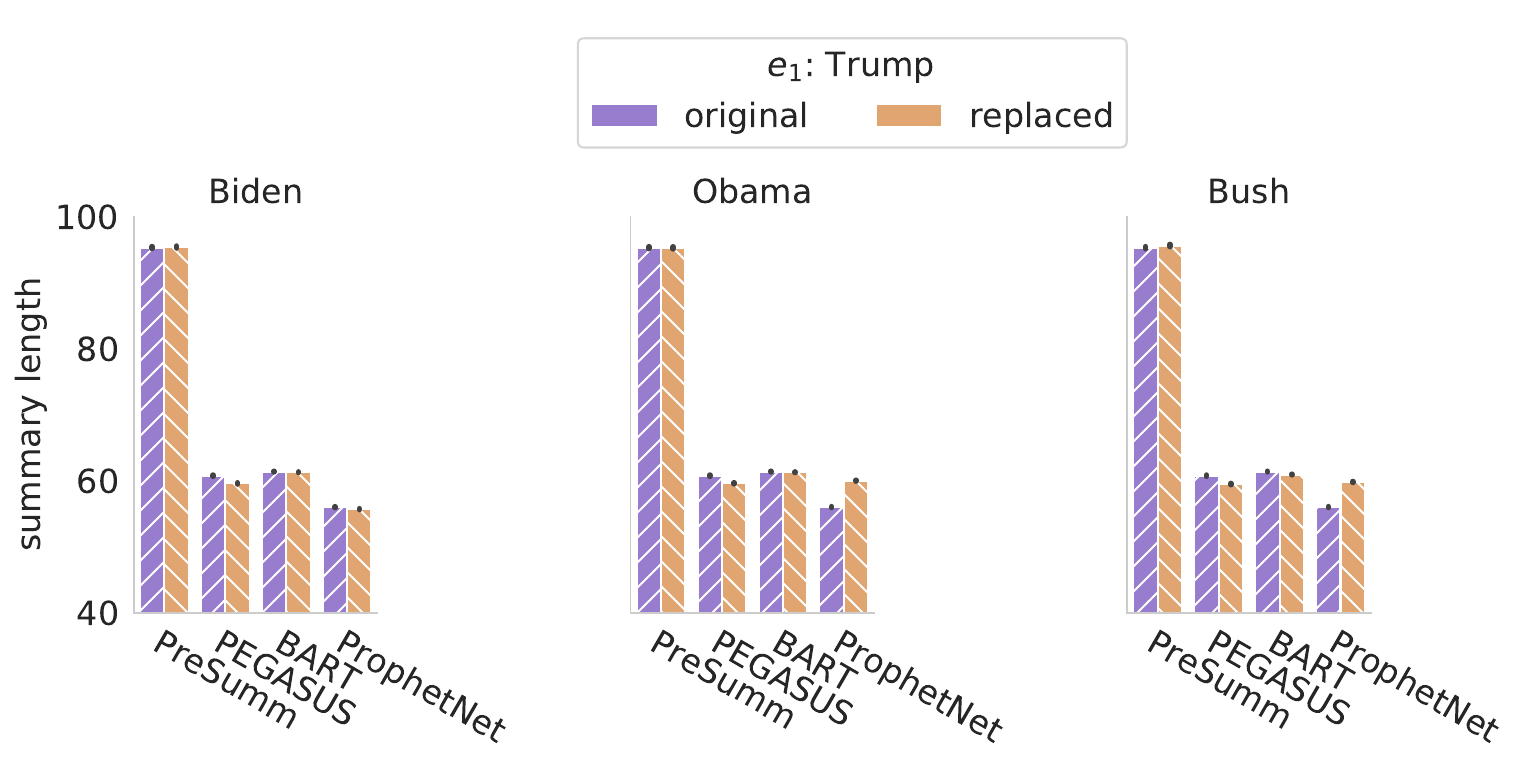}
		\includegraphics[width=0.35\textwidth]{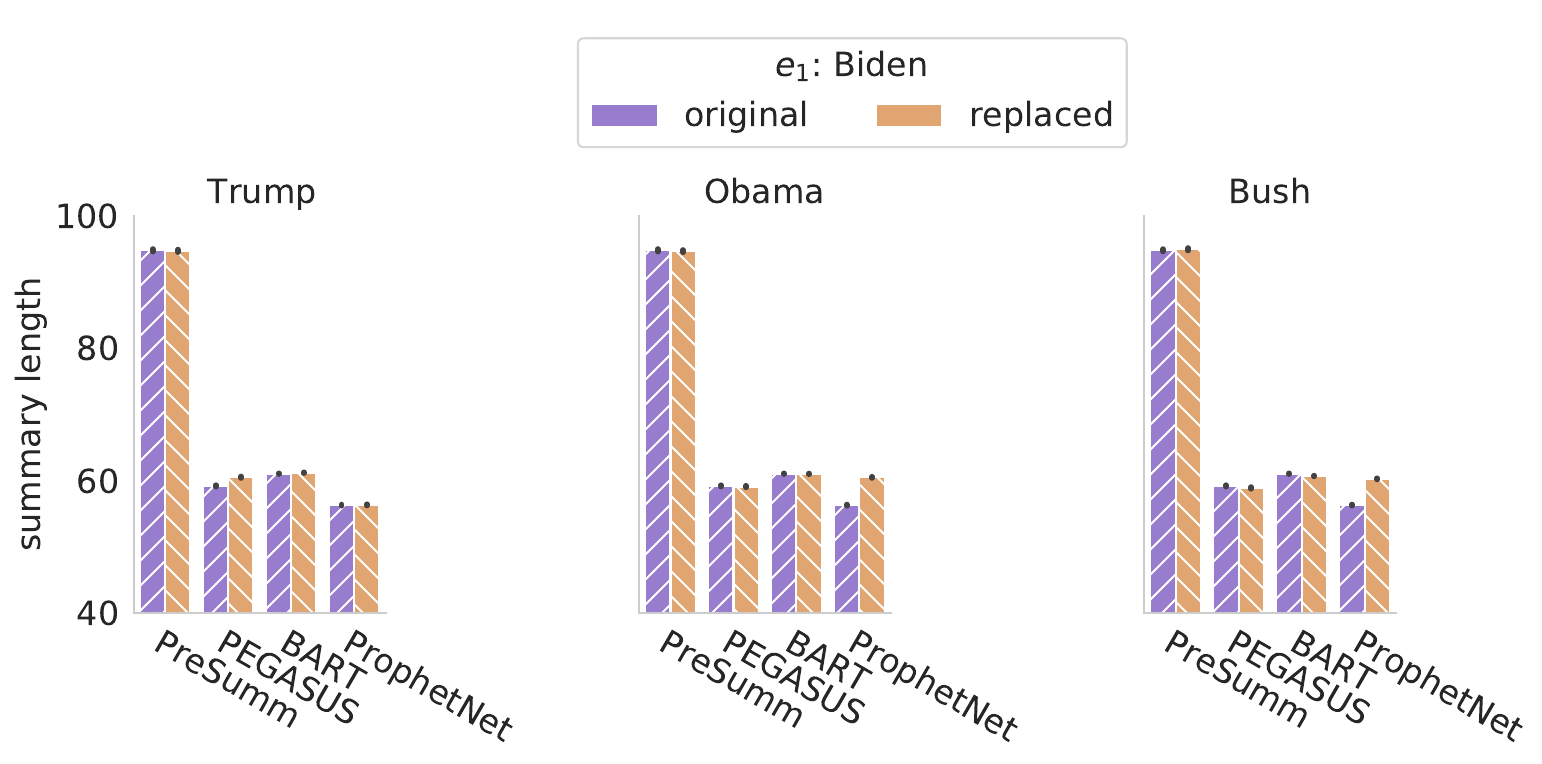}

	\caption{There are small but significant differences in length between the original and replaced summaries.
	}
	\label{fig:sum_lens}
\end{figure}

\section{Additional Summary Similarity Comparisons}
\label{sec:app_sim_comps}

\para{Comparing lengths and similarity scores of $S_i^{e_1}$ and $S_i^{e_2}$:}
\label{ssec:lens}
To obtain lengths of summaries, we count the number of tokens as parsed by spaCy's tokenizer.
There are small but significant differences between the lengths of $S_i^{e_1}$ and $S_i^{e_2}$, as shown in \figref{fig:sum_lens}.
From a paired $t$-test, we find that abstractive summaries tend to be longer for Trump (see \tbref{tab:app_sum_ttest}). 
Note that 
\textsc{PreSumm} summaries are longer due to default model settings. Again, we control for these discrepancies by normalizing all feature function frequencies by summary length.

\begin{figure*}
	\centering
	\begin{subfigure}[t]{0.45\textwidth}
		\centering
		\includegraphics[width=\textwidth]{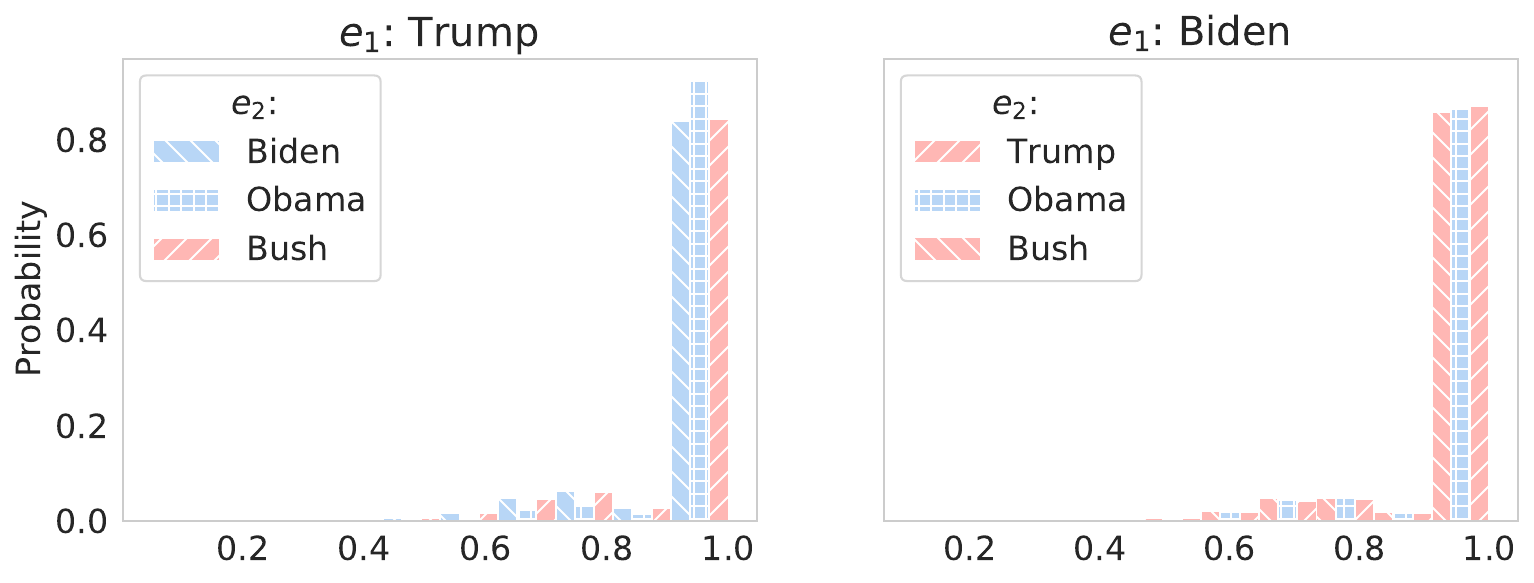}
		\caption{\PreSumm}
		\label{fig:}
	\end{subfigure}
	\begin{subfigure}[t]{0.45\textwidth}
		\centering
		\includegraphics[width=\textwidth]{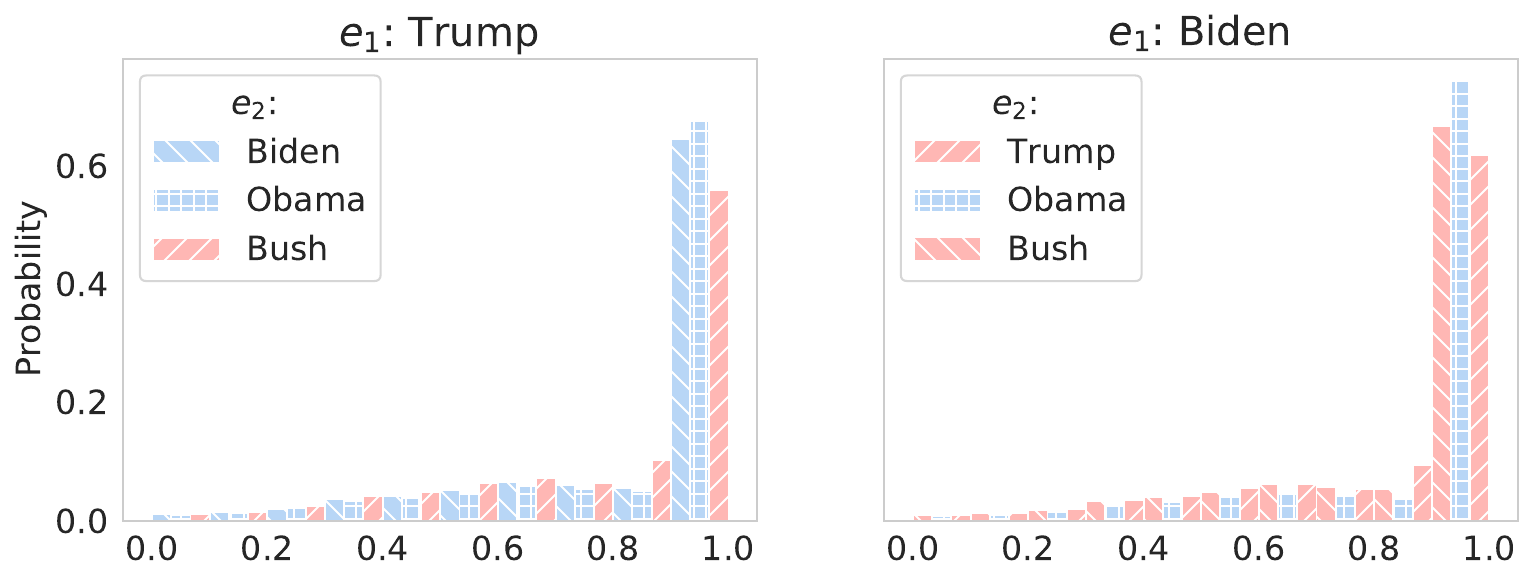}
		\caption{\PEGASUS}
		\label{fig:}
	\end{subfigure}
	
	\bigskip
	\medskip
	\begin{subfigure}[t]{0.45\textwidth}
		\centering
		\includegraphics[width=\textwidth]{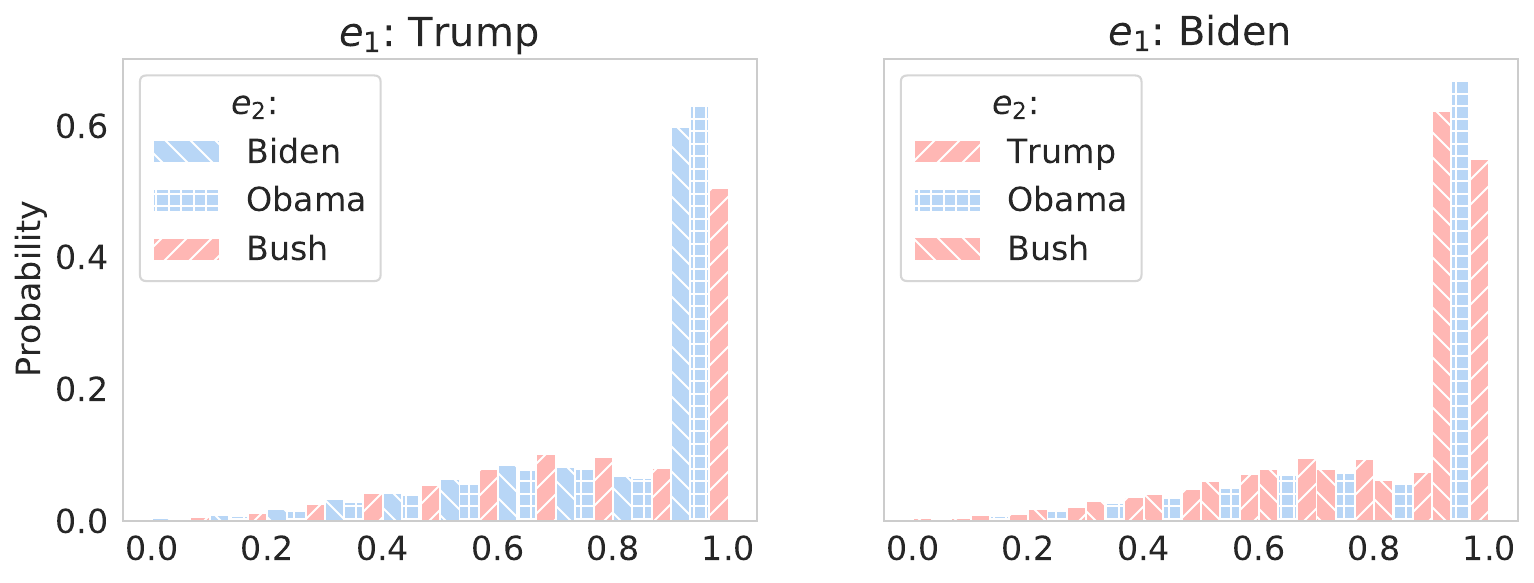}
		\caption{\BART}
		\label{fig:}
	\end{subfigure} 
	\begin{subfigure}[t]{0.45\textwidth}
		\centering
		\includegraphics[width=\textwidth]{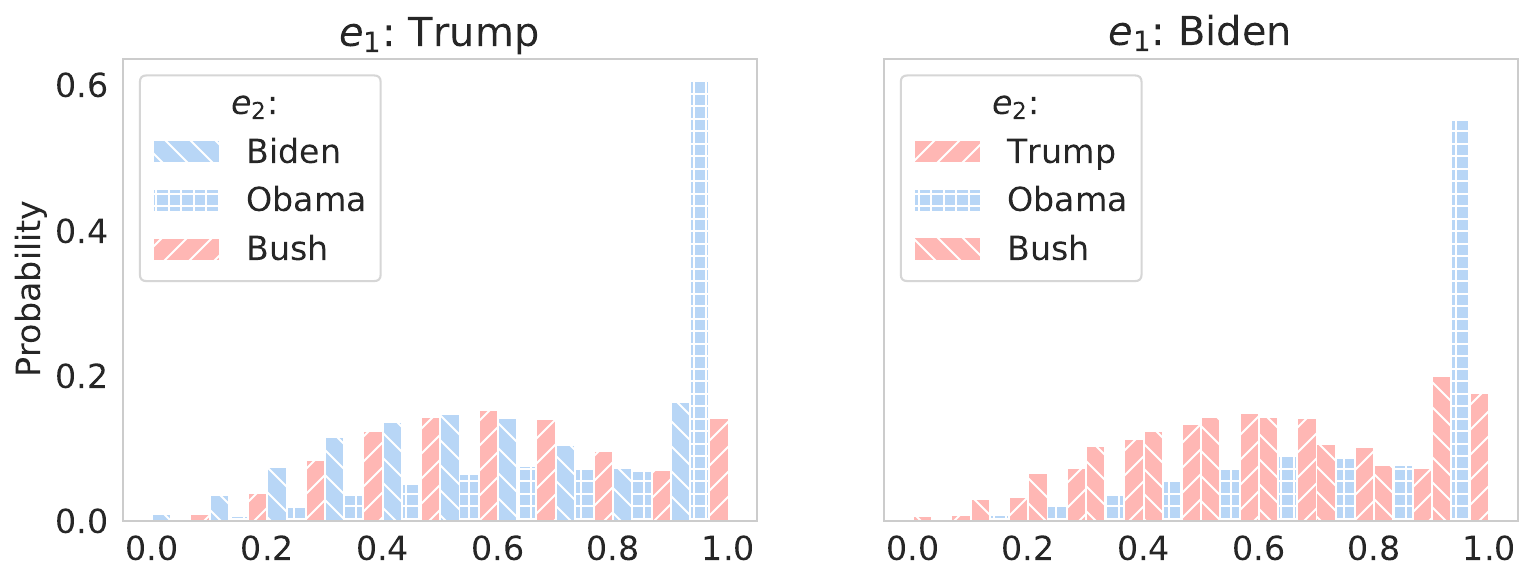}
		\caption{\ProphetNet}
		\label{fig:}
	\end{subfigure}
	\caption{Distribution of similarity scores across models. \ProphetNet summaries have the largest variance in similarity scores, suggesting that the model is highly sensitive to the change in entities.
	}
	\label{fig:sim_dists}
\end{figure*}

The distributions of similarity scores for \TB  and \BT  summaries are shown in \figref{fig:sim_dists}.  \ProphetNet summaries have the largest variance in similarity scores, suggesting that it is highly sensitive to the change in entities.

\begin{table*}
	\small
	\centering
	\setlength\extrarowheight{2pt} 
	\begin{tabular}{ p{1.5cm} p{6.75cm} p{6.75cm} }
	\toprule
	\textbf{Model} & \textbf{Trump $\to$ Biden} & \textbf{Biden $\to$ Trump}\\
	\midrule
	\PreSumm
	& \ctext[RGB]{139,204,198}{county (31.365), city (26.906), school (25.667), she (25.442), her (25.396), year (24.547), home (24.080), cases (23.626), covid19 (22.925), season (22.865)}, 
	\ctext[RGB]{247,198,80}{trump (-256.244), president (-136.788), house (-79.619), donald (-72.599), impeachment (-66.696), his (-62.088), white (-57.215), election (-54.234), he (-54.205), republican (-50.200)}	
	&
	\ctext[RGB]{139,204,198}{company (28.989), its (23.543), city (22.786), year (22.246), market (21.871), police (19.597), was (18.324), cases (18.192), in (17.913), game (17.546)}, 
\ctext[RGB]{247,198,80}{biden (-156.466), president (-75.908), house (-56.627), democrats (-51.874), joe (-50.807), infrastructure (-49.757), white (-46.655), administration (-44.411), bill (-41.257), senate (-40.555)}
	\\
	\midrule
	\PEGASUS  
	& \ctext[RGB]{139,204,198}{her (41.645), county (37.444), year (37.189), she (36.660), city (30.585), 2021 (30.479), 2020 (30.161), police (28.710), film (27.468), market (27.268)}, 
	\ctext[RGB]{247,198,80}{trump (-307.278), president (-155.661), house (-92.642), he (-80.623), his (-76.141), white (-75.438), donald (-68.320), administration (-68.019), impeachment (-61.816), republican (-53.615)}	
	&
	\ctext[RGB]{139,204,198}{market (39.070), company (38.515), year (32.385), 2021 (32.382), its (31.485), stock (30.825), shares (27.808), investors (27.771), city (26.918), stocks (26.547)}, 
\ctext[RGB]{247,198,80}{biden (-188.517), president (-82.308), house (-74.851), administration (-71.761), white (-65.998), democrats (-60.348), americans (-50.569), infrastructure (-50.390), republicans (-47.779), joe (-47.057)}
	\\
	\hline
	\BART 
	& \ctext[RGB]{139,204,198}{her (39.586), year (38.537), she (38.304), county (36.911), city (32.461), 2021 (31.370), film (30.693), march (30.361), school (29.533), season (29.488)}, 
	\ctext[RGB]{247,198,80}{trump (-325.481), president (-170.838), house (-102.056), white (-87.634), administration (-79.610), he (-79.251), his (-75.671), donald (-72.398), impeachment (-68.006), election (-58.328)}
	&
	\ctext[RGB]{139,204,198}{company (45.275), market (37.283), you (34.736), 2021 (33.281), year (31.927), investors (30.914), shares (30.718), city (30.488), stock (30.376), its (29.238)}, 
\ctext[RGB]{247,198,80}{biden (-204.154), president (-89.259), administration (-83.676), house (-83.446), white (-76.504), democrats (-67.607), infrastructure (-58.296), americans (-55.697), senate (-53.701), republicans (-52.966)}
	 \\
	\hline
	\ProphetNet  
	& \ctext[RGB]{139,204,198}{said (46.719), china (32.564), reuters (30.071), us (29.690), coronavirus (29.533), its (29.109), in (28.481), company (26.431), cases (25.838), 2020 (25.116)}, 
	\ctext[RGB]{247,198,80}{trump (-93.315), you (-54.985), nt (-41.435), his (-38.105), he (-35.592), do (-34.547), it (-33.174), impeachment (-33.131), that (-32.700), if (-32.344)}	
	&
	\ctext[RGB]{139,204,198}{reuters (29.241), 2021 (28.519), its (28.260), company (25.412), market (24.053), shares (22.754), said (22.629), stocks (20.942), china (20.300), apple (19.893)}, 
\ctext[RGB]{247,198,80}{biden (-54.498), nt (-30.933), that (-28.089), we (-26.924), you (-26.287), they (-24.591), do (-24.516), his (-24.345), he (-23.782), what (-23.561)}
	\\
	\bottomrule
	\end{tabular}
\caption{Full set of key words, with Fightin' words z-score in parentheses \citep{monroe_fightin_2009}, for \textsc{HighSim} and \textsc{LowSim} articles.
\ctext[RGB]{139,204,198}{Teal highlight} corresponds to \textsc{HighSim}, while \ctext[RGB]{247,198,80}{gold highlight} corresponds to \textsc{LowSim}.
}
\label{tab:bart_prophetnet_highsim_low_sim}
\end{table*}

\para{Factors driving summary dissimilarity:}
\label{ssec:app_factors}
Table~\ref{tab:bart_prophetnet_highsim_low_sim} shows the the differences between key words in \textsc{HighSim} and \textsc{LowSim} articles for all models.  Note the high occurrence of {\em entity-centric} names and words for \textsc{LowSim} articles.

``Impeachment'' articles leads to low similarity summaries for \TB; ``infrastructure'' 
 articles likewise lead to dissimilar articles for \BT. 
Meanwhile, key words ``coronavirus''/ ``COVID-19'' and ``police''
 correspond to \textsc{HighSim}.
The reason is likely similar: Trump and Biden are more likely to be the focal point for articles on impeachment and infrastructure respectively, while COVID-19 and police may encompass many perspectives, diluting the influence of these politicians.

\input{fighting_words.tex}

\input{examples.tex}

%% file: fighting_words.tex
\para{Fightin' words of $S_i^{e_1}$ and $S_i^{e_2}$ and hallucinations:} 
We use the Fightin’ Words algorithm \citep{monroe_fightin_2009} on the original vs. replaced summaries.  \tbref{fighting_words} shows the results for \TB  and \BT  replacements.

There is strong evidence of hallucination among the abstractive models, particularly of journalist and reporter names (e.g., ``Ruben Navarrette''). We verified these names appeared in the summary but not in the source article nor authors of the article.
For example, this \PEGASUS summary was generated for this Trump only \href{https://www.dispatch.com/story/opinion/editorials/2020/06/22/editorial-daca-decision-is-clear-congress-must-fix-this/42156271/}{Columbus Dispatch article} (that includes no mention of Navarrette): 

\begingroup
\advance\leftmargini -1.5em
\begin{quote}
	\emph{The Supreme Court shot down the Trump administration's efforts to undo the Obama-era Deferred Action for Childhood Arrivals program. Ruben Navarrette: DACA is a temporary solution to a broader problem. He says Congress must get its act together and take an obvious step in the national interest. Navarrette: An overwhelming majority of Americans. believe the government should leave the Dreamers alone and craft a path to citizenship.}
\end{quote}
\endgroup

\begin{table*}
\small
	\centering
	\setlength\extrarowheight{2pt} 
	\begin{tabular}{ m{1.75cm}  p{6.5cm} p{6.5cm} }
	\toprule
	\textbf{Model} & \textbf{Trump $\to$ Biden} & \textbf{Biden $\to$ Trump}\\
	\hline
	\PreSumm
	& \ctext[RGB]{218,191,229}{trumpian (2.673),     (2.629), trumped (1.910), trumpist (1.710), preys (1.659), realdonaldtrump (1.554), strumpf (1.469), administration (1.459), trumpists (1.445), question (1.396)}, 
	\ctext[RGB]{248,182,105}{donald (-7.731), president (-3.663), trump (-2.644), his (-2.165), he (-2.157), impeached (-1.946), repeatedly (-1.776), pazjune (-1.467), escalator (-1.439), inciting (-1.428)}
	&
	\ctext[RGB]{218,191,229}{joseph (7.555), joe (3.047), president (2.202), his (1.893), trumpeted (1.751), he (1.584), curtailed (1.401), elect (1.200), decreed (1.192), instantaneous (1.129)}, 
	\ctext[RGB]{248,182,105}{bidenism (-2.678), administration (-2.213), bidens (-1.281), 1of2a (-1.239), wantagh (-1.239), 11475 (-1.095), farrar (-1.095),     
	langevin (-0.991), 17171 (-0.981)}
	\\
	\hline
	\PEGASUS 
	& \ctext[RGB]{218,191,229}{but (6.458), kohn (5.129), shoulders (4.750), president (4.611), squarely (4.550), it (4.548), said (4.462), deere (3.437), experts (3.368), the (3.258)}, 
	\ctext[RGB]{248,182,105}{trump (-37.403), administration (-12.995), vice (-11.207), has (-5.742), rescuers (-5.464), his (-4.993), debris (-4.314), campaign (-4.242), obama (-4.138), supporters (-3.933)}
	&
	\ctext[RGB]{218,191,229}{biden (30.578), administration (13.626), vice (5.527), joseph (4.611), his (4.230), has (3.913), plan (3.097), says (2.847), announce (2.430), bidens (2.343)}, 
	\ctext[RGB]{248,182,105}{kohn (-4.040), but (-3.776), it (-3.129), white (-3.061), house (-2.944), experts (-2.605), there (-2.540), that (-2.538), president (-2.529), anonymity (-2.398)}
	\\
	\hline
	\BART 
	& \ctext[RGB]{218,191,229}{president (9.808), the (8.552), wellington (5.263), navarrette (5.214), ruben (4.985), bergen (4.681), didt (3.921), mr (3.733), trumpian (3.669), house (3.506)}, 
	\ctext[RGB]{248,182,105}{trump (-20.222), administration (-10.621), avlon (-6.675), vice (-4.694), heavycom (-3.500), diana (-3.413), prince (-3.337), reunite (-3.332), princess (-3.181), fabled (-3.041)}
	&
	\ctext[RGB]{218,191,229}{biden (19.605), administration (11.651), joseph (5.858), avlon (3.495), zelizer (2.687), says (2.329), incompetence (2.277), plan (2.223), joebiden (1.981), obeido (1.964)}, 
	\ctext[RGB]{248,182,105}{the (-6.473), white (-5.956), president (-5.813), house (-4.566), mr (-3.589), navarrette (-3.362), ruben (-3.356), brazile (-2.559), lz (-2.377), sutter (-2.180)}
	 \\
	\hline
	\ProphetNet  
	& \ctext[RGB]{218,191,229}{kohn (16.515), sally (9.127), said (8.751), president (7.736), the (7.385), noisy (6.977), sutter (5.895), donald (5.770), cnn (4.515), gervais (3.892)}, 
	\ctext[RGB]{248,182,105}{obama (-14.729), frum (-14.633), administration (-10.525), trump (-8.880), david (-6.602), brazile (-5.914), former (-5.631), says (-5.568), obeidallah (-5.330), vp (-5.196)}
	&
	\ctext[RGB]{218,191,229}{administration (10.932), biden (10.800), frum (8.866), obama (8.352), joseph (4.936), david (3.925), brazile (3.797), says (3.796), carroll (3.595), cdc (3.536)}, 
	\ctext[RGB]{248,182,105}{joe (-11.879), president (-11.014), kohn (-9.294), the (-8.259), said (-6.787), white (-5.547), sally (-5.445), namibia (-5.275), house (-4.849), wal (-3.741)}
	\\
	\bottomrule
	\end{tabular}
\caption{\label{fighting_words}
Fightin' words for $S_i^{e_1}$ vs. $S_i^{e_2}$, with z-score in parentheses \citep{monroe_fightin_2009}.
\ctext[RGB]{218,191,229}{Purple highlight} corresponds to the original summaries $S_i^{e_1}$ , while \ctext[RGB]{248,182,105}{orange highlight} corresponds to the 
replaced summaries $S_i^{e_2}$ 
}
\end{table*}

%% file: examples.tex
\newcommand{\orig}[1]{\textcolor{violet}{#1}}
\newcommand{\repl}[1]{\textcolor{orange}{#1}}
\newcommand{\source}[1]{\href{#1}{source}}

\section{Notable Examples}
\label{app:ex}

The following examples are for the replacements \TB and \BT only. 

\para{Entity name frequency} \tbref{tab:ex_freqs} shows examples of summaries that differ in frequency of entity name mentions; in particular, Trump's name is de-emphasized compared to Biden's.

\para{``Vice President''} \tbref{tab:ex_vp} shows examples of summaries that differ in usage of ``Vice President'', in particular associating the title with Joe Biden.

\input{example_table.tex}

\para{``Administration''}  \tbref{tab:ex_admin} shows examples of summaries that differ in usage of ``administration'', in particular in association with Joe Biden.

%% file: example_table.tex
\begin{table*}
	\footnotesize
		\centering
		\setlength\extrarowheight{2pt} 
		\begin{tabular}{ p{2.75cm}  p{6cm} p{6cm}}
		\toprule
		\textbf{Model, Instantiation} & \textbf{
			Original Summary
			} & \textbf{Replaced Summary}\\
		\hline
		\BART
		\newline
		\TB
		\newline
		\source{https://www.huffpost.com/entry/tony-schwartz-trump-out-of-control_n_5f629236c5b6c6317cff40ea}
		& Tony Schwartz helped write ``\orig{Trump} : The Art of the Deal'' He says he's never seen the president more frightening, out of control and disconnected with reality than he is now. Schwartz: ``We are in a relentless gaslighting in which he does lie and deceive multiple times a day''
		& Tony Schwartz helped write ``\repl{Biden} : The Art of the Deal'' He says \repl{Biden} is ``in full gaslighting mode \& willing to say anything to survive'' He attributes \repl{Biden}'s absence of conscience and empathy to sociopathy. Schwartz says social media companies should respond to \repl{Biden}'s peddling of manipulated content.
		\\
		\hline
		\PEGASUS
		\newline
		\BT
		\newline
		\source{https://news.trust.org/item/20210728130018-ghpad}
		& The White House is signaling to U.S. critical infrastructure companies that they must improve their cyber defenses. U.S. President \orig{Joseph Biden} signed a national security memorandum on Wednesday. \orig{Biden} warned that if the United States ended up in a ``real shooting war " with a ``major power" it could be the result of a significant cyber attack on the United States.
		& The White House is signaling to U.S. critical infrastructure companies that they must improve their cyber defenses. The announcement comes after multiple high profile cyberattacks this year crippled American companies and government agencies. Almost 90\% of critical infrastructure is owned and operated by the private sector. 
		\\
		\hline
		\ProphetNet  \newline \TB \newline \source{https://abcnews.go.com/Politics/trump-republicans-asked-states-stop-counting-votes/story?id=74043767}
		& president \orig{donald trump} has called for vote counting to stop in the 2018 midterm elections.  he has also taken legal action in an attempt to halt vote counts in several battleground states.  the president's comments have sparked ``count the vote" protest marches across the country.
		& \repl{biden} campaign has taken legal action in an attempt to halt vote counts in several battleground states.  \repl{biden} supporters have gathered in both states, calling to ``stop the count"  \repl{biden}'s comments have sparked protest marches across the country.
		\\
		\bottomrule
		\end{tabular}
	\caption{\label{tab:ex_freqs}
		Examples of summaries demonstrating the findings about usage of entity names in summaries.
	}
\end{table*}

\begin{table*}
	\footnotesize
		\centering
		\setlength\extrarowheight{2pt} 
		\begin{tabular}{ p{3cm}  p{5.5cm} p{5.5cm}}
		\toprule
		\textbf{Model, Instantiation} & \textbf{
			Original Summary
			} & \textbf{Replaced Summary}\\
		\hline
		\PreSumm \newline \BT \newline 
		\source{https://www.scpr.org/news/2021/01/23/96306/opinion-joe-biden-s-lifetime-of-experience/} 
		& A year ago, who would have thought 78-year-old \orig{Joe Biden} would be sworn in this week as president? His wife and infant daughter died in a car accident in 1972, just weeks before he was sworn in as a U.S. senator. He served two terms as Barack Obama's \orig{vice president}, but after \orig{Biden} didn't run again in 2016, was widely thought to be past his expiration date in public office. 
		& A year ago, who would have thought 78-year-old \repl{Donald Trump} would be sworn in this week as president? His wife and infant daughter died in a car accident in 1972, just weeks before he was sworn in as a U.S. senator.  Many of the 81 million people who voted for \repl{Donald Trump} and Kamala Harris probably know about the president's gaffes, mistakes and flaws. 
		\\
		\hline
		\PEGASUS
		\newline
		\TB
		\newline
		\source{https://www.engadget.com/donald-trump-snapchat-ban-004602436.html}
		& Snapchat has permanently suspended \orig{President Trump}'s account. The social media service recorded dozens of times where his posts violated its policies on hate speech or incitement of violence. Twitter, Twitch, Facebook and others no longer offer access to Donald Trump. 
		& \repl{Vice President Joe Biden}'s Snapchat account has been suspended indefinitely. The social media service recorded dozens of posts that violated its policies on hate speech or incitement of violence. Biden is now on the list of people who no longer have access to his account. 
		 \\
		\hline
		\PEGASUS
		\newline
		\BT
		\newline
		\source{https://www.tmz.com/2021/01/21/amanda-gorman-books-amazon-best-sellers-poet-laureate-inauguration-joe-biden/}
		& Amanda Gorman is the youngest inaugural poet in U.S. history. Her poem at the Inauguration was read by \orig{Vice President Joe Biden}. Gorman's books are already No. 1 and No. 2 on Amazon.
		& Amanda Gorman is the youngest inaugural poet in U.S. history. Her poem at \repl{President Trump}'s Inauguration was read by first lady Michelle Obama. Gorman's books are already No. 1 and No. 2 on Amazon. 
		\\
		\hline
		\PEGASUS
		\newline
		\BT
		\newline
		\source{https://www.thedailybeast.com/biden-postpones-sunday-inauguration-rehearsal-over-security-threats}
		& \orig{Vice President-elect Joe Biden} has canceled an inauguration rehearsal scheduled for Sunday over threats he and his team have received. The official inauguration is slated for Wednesday. Biden had also scheduled to take an Amtrak train from Wilmington, Delaware to the capital.
		& \repl{President-elect Donald Trump} has canceled an inauguration rehearsal scheduled for Sunday over threats he and his team have received. The official inauguration is slated for Wednesday. The Jan. 6 riot in the U.S. Capitol building has heightened security concerns in Washington. 
		\\
		\hline
		\BART  \newline \TB \newline \source{https://thehill.com/opinion/campaign/536669-juan-williams-gop-cowers-from-qanon}
		& Marjorie Taylor Greene, a Republican in Georgia, is a supporter of QAnon, a conspiracy theory. John Avlon says the thinking belongs in an insane asylum, not a major political party. Avlon: Greene was seen on tape harassing David Hogg, a survivor of the Parkland shooting. He says GOP leadership should expel Greene from Congress.
		& Marjorie Taylor Greene was endorsed by \repl{Vice President Joe Biden}. John Avlon: Greene has supported the conspiracy theory that the shooting at Hogg's school was staged. Avlon says GOP leadership should have expelled Greene from Congress. GOP strategists might fear losing her supporters and the whole QAnon crowd in 2022 elections. 
		\\
		\hline
		\BART \newline \BT \newline \source{https://thehill.com/homenews/administration/555427-biden-travels-to-delaware-to-pay-respects-to-family-of-longtime} 
		& Norma Long worked for \orig{Vice President Joe Biden} for more than 30 years. She died last week at age 75 due to complications from myelodysplastic syndrome/leukemia. Head of DHS says he expects ''significant changes'' to U.S. Immigration and Customs Enforcement. 
		& The \repl{president} made a quick stop in Delaware to pay respects to the family of Norma Long. Long died last week at age 75 due to complications from myelodysplastic syndrome/leukemia. Long worked for \repl{Trump} in 1977 for his reelection campaign. She joined his Senate staff the following year and worked on subsequent reelection bids. 
		\\
		\hline
		\ProphetNet  \newline \TB \newline \source{https://www.brookings.edu/blog/fixgov/2021/02/05/trumps-future-nine-possibilities/}
		& timothy stanley: \orig{trump}'s second impeachment trial shows he is not going away.  he says the first four scenarios keep him in the middle of national politics.  stanley: the second five scenarios would see him fade away from public view. he says the most likely scenario is that \orig{trump} will continue to be the party's leader. 
		& timothy stanley: \repl{joe biden}'s second impeachment trial is a new reality.  he says \repl{biden}'s ''\repl{bidenublican}'' faction of gop could continue to influence gop. he says \repl{biden} could be a strong candidate in 2016 but fade away if he loses. stanley: \repl{biden} could be a moderate or a strong candidate for \repl{vice president}. 
		\\
		\hline
		\ProphetNet  \newline \BT \newline \source{https://www.foxnews.com/politics/man-arrested-in-north-carolina-had-van-full-of-guns-explosives-planned-to-assassinate-joe-biden-report}
		& alexander hillel treisman is accused of plotting to kill \orig{vice president joe biden}. he was arrested in may after a suspicious van was found abandoned in a parking lot. a federal judge said treisman's alleged plot against \orig{biden} is the reason for his jail time. he is also charged with child pornography charges. 
		& alexander hillel treisman was arrested in late may.  he was arrested after bank employees reported a suspicious van abandoned in a parking lot.  treisman is accused of plotting to kill \repl{donald trump}.  he is being held on child pornography charges. 
		\\
		\bottomrule
		\end{tabular}
	\caption{\label{tab:ex_vp}
		Examples of summaries demonstrating the findings about usage of ``Vice President''.
	}
\end{table*}

\begin{table*}
	\footnotesize
		\centering
		\setlength\extrarowheight{2pt} 
		\begin{tabular}{ p{3cm}  p{5.5cm} p{5.5cm} }
		\toprule
		\textbf{Model, Instantiation} & \textbf{
			Original Summary
			} & \textbf{Replaced Summary}\\
		\hline
		\PEGASUS \newline \TB \newline \source{https://www.thedailybeast.com/they-have-covid-one-has-an-iq-of-69-trump-is-killing-them}  
		& At least 309 inmates and nearly 30 staff at the federal prison complex in Terre Haute, Indiana have active COVID cases. Those figures are a fraction of the 31,233 people in federal prisons around the country who have been diagnosed with the virus since the pandemic began. More than 6,000 federal inmates are currently ill with COVID.  163 others have died. 
		& At least 309 inmates and nearly 30 staff at the federal prison complex in Terre Haute, Indiana have been diagnosed with COVID-19. Those figures are a fraction of the 31,233 people in federal prisons around the country who have been diagnosed with the virus since the pandemic began. The government's failure to protect prisoners from COVID, and the inmate deaths from the disease, should be regarded as a result of the same callous indifference to human life that spurred the \repl{Biden administration} to. 
		\\
		\hline
		\PEGASUS \newline \BT \newline \source{https://www.fool.com/the-ascent/personal-finance/articles/biden-to-reveal-high-priced-stimulus-plan-this-week-with-2000-checks/}  
		& \orig{Biden} indicated he plans to lay out his proposed aid package in detail this week -- and that it will include \$2,000 stimulus checks. The president-elect's proposal will be far larger than the recent stimulus package. The \orig{administration} has already begun talks with Democrats on a proposed bill in hopes of quickly moving a relief any legislation will need 60 votes in the Senate to pass.
		& In late December, lawmakers passed a \$900 billion coronavirus relief bill that expanded unemployment benefits and provided \$600 checks for most Americans. With Democrats winning control of the U.S. Senate, it's almost certain more coronavirus stimulus aid will be a top priority. In fact, \repl{Trump} indicated he plans to lay out his proposed aid package in detail this week -- and that it will include \$2,000 stimulus checks.
		\\
		\hline
		\BART \newline \TB \newline \source{https://www.tallahassee.com/story/opinion/2021/01/19/capitol-attack-both-disrespect-trump-and-trump-himself-blame/4202094001/}
		& Kevin Burns: The Capitol Hill tragedy was like an aircraft accident. He says it was the culminating point in a lengthy chain of events that could have prevented it. Burns: Both disrespect of \orig{Trump} and \orig{Trump} himself to blame for attack on Capitol. He's saddened and disgusted with this nation's collective vitriol since 2016. 
		& Kevin Burns: Jan. 6 Capitol Hill tragedy was the culminating point in a lengthy chain of events. He says many of the chain events revolved around the unrelenting disrespect heaped on the current \repl{administration}. Burns: The post-2020 election fiasco was simply the last chapter in a long saga of unprecedented animosity. 
		\\
		\hline
		\BART \newline \BT \newline \source{https://thehill.com/opinion/energy-environment/530967-reentering-the-paris-climate-agreement-is-a-must-for-the-us-but}
		& This month marks the fifth anniversary of the Paris Climate Agreement. Jonathan Jennings: The \orig{Biden administration} appears poised to rejoin the multilateral pact. Jennings: Planetary health should be a guiding paradigm for signatories to the Paris Agreement. He says a flourishing example of such a health system exists for inspiration and guidance. 
		& This month marks the fifth anniversary of the Paris Climate Agreement. Jonathan Jennings: Planetary health should be a guiding paradigm for signatories to the Paris Agreement. Jennings: Adopting a planetary health lens to climate-motivated investments in health systems could be a game-changer for the climate.
		\\
		\hline
		\ProphetNet \newline \TB \newline \source{https://www.businessinsider.com/taiwan-says-asking-chip-firms-to-help-ease-auto-chip-shortage-2021-1}
		& automakers around the world are shutting assembly lines because of a global shortage of semiconductors.  automakers around the world are shutting assembly lines because of a global shortage of semiconductors.  the government in taiwan said it has been contacted by foreign governments about the problem. 
		& automakers around the world are shutting assembly lines because of a global shortage of semiconductors.  the shortage has been exacerbated by the former \repl{biden administration}'s actions against key chinese chip factories.  taiwan's ministry of economic affairs has asked local tech firms to provide  ``full assistance '' 
		\\
		\hline
		\ProphetNet \newline \BT \newline \source{https://www.wxyz.com/news/national/coronavirus/president-elect-joe-biden-to-receive-second-dose-of-pfizer-vaccine-today}
		& \orig{biden} received his second covid - 19 vaccine on monday.  he received the injection in his home state of delaware.  \orig{biden's administration} wants to distribute 100 million doses of the vaccine. 
		& president - elect \repl{donald trump} got his second covid - 19 vaccine on monday.  \repl{trump} received the first dose of the vaccine on dec. 21.  \repl{the u.s.} has distributed 8 million vaccine doses since dec. 14. 
		\\
		\bottomrule
		\end{tabular}
	\caption{\label{tab:ex_admin}
	Examples of summaries demonstrating the findings about usage of ``administration''.
	}
\end{table*}

%% file: main.bbl
\begin{thebibliography}{45}
\expandafter\ifx\csname natexlab\endcsname\relax\def\natexlab#1{#1}\fi

\bibitem[{Bagdasaryan and Shmatikov(2021)}]{bagdasaryan_spinning_2021}
Eugene Bagdasaryan and Vitaly Shmatikov. 2021.
\newblock \href {https://doi.org/10.48550/arXiv.2112.05224} {Spinning
  {{Language Models}}: {{Risks}} of {{Propaganda-As-A-Service}} and
  {{Countermeasures}}}.

\bibitem[{Buchanan et~al.(2021)Buchanan, Lohn, Musser, and
  Sedova}]{buchanan_truth_2021}
Ben Buchanan, Andrew Lohn, Micah Musser, and Katerina Sedova. 2021.
\newblock Truth, {{Lies}}, and {{Automation}}.

\bibitem[{Chong and Druckman(2007)}]{chong2007theory}
Dennis Chong and James~N Druckman. 2007.
\newblock A theory of framing and opinion formation in competitive elite
  environments.
\newblock \emph{Journal of communication}, 57(1):99--118.

\bibitem[{Devaraj et~al.(2022)Devaraj, Sheffield, Wallace, and
  Li}]{devaraj_evaluating_2022}
Ashwin Devaraj, William Sheffield, Byron Wallace, and Junyi~Jessy Li. 2022.
\newblock Evaluating {{Factuality}} in {{Text Simplification}}.
\newblock In \emph{Proceedings of the 60th {{Annual Meeting}} of the
  {{Association}} for {{Computational Linguistics}} ({{Volume}} 1: {{Long
  Papers}})}, pages 7331--7345, {Dublin, Ireland}. {Association for
  Computational Linguistics}.

\bibitem[{Dhamala et~al.(2021)Dhamala, Sun, Kumar, Krishna, Pruksachatkun,
  Chang, and Gupta}]{dhamala_bold_2021}
J.~Dhamala, Tony Sun, Varun Kumar, Satyapriya Krishna, Yada Pruksachatkun,
  Kai-Wei Chang, and Rahul Gupta. 2021.
\newblock \href {https://doi.org/10.1145/3442188.3445924} {{{BOLD}}:
  {{Dataset}} and {{Metrics}} for {{Measuring Biases}} in {{Open-Ended Language
  Generation}}}.
\newblock \emph{FAccT}.

\bibitem[{{El-Kassas} et~al.(2021){El-Kassas}, Salama, Rafea, and
  Mohamed}]{el-kassas_automatic_2021}
Wafaa~S. {El-Kassas}, Cherif~R. Salama, Ahmed~A. Rafea, and Hoda~K. Mohamed.
  2021.
\newblock \href {https://doi.org/10.1016/j.eswa.2020.113679} {Automatic text
  summarization: {{A}} comprehensive survey}.
\newblock \emph{Expert Systems with Applications}, 165:113679.

\bibitem[{Entman(1993)}]{entman1993framing}
Robert~M Entman. 1993.
\newblock Framing: Towards clarification of a fractured paradigm.
\newblock \emph{McQuail's reader in mass communication theory}, pages 390--397.

\bibitem[{Feder et~al.(2021)Feder, Oved, Shalit, and
  Reichart}]{feder_causalm_2021}
Amir Feder, Nadav Oved, Uri Shalit, and Roi Reichart. 2021.
\newblock \href {https://doi.org/10.1162/coli_a_00404} {{{CausaLM}}: {{Causal
  Model Explanation Through Counterfactual Language Models}}}.
\newblock \emph{Computational Linguistics}, pages 1--54.

\bibitem[{Feng et~al.(2023)Feng, Park, Liu, and
  Tsvetkov}]{feng-etal-2023-pretraining}
Shangbin Feng, Chan~Young Park, Yuhan Liu, and Yulia Tsvetkov. 2023.
\newblock \href {https://doi.org/10.18653/v1/2023.acl-long.656} {From
  pretraining data to language models to downstream tasks: Tracking the trails
  of political biases leading to unfair {NLP} models}.
\newblock In \emph{Proceedings of the 61st Annual Meeting of the Association
  for Computational Linguistics (Volume 1: Long Papers)}, pages 11737--11762,
  Toronto, Canada. Association for Computational Linguistics.

\bibitem[{Garimella et~al.(2021)Garimella, Amarnath, Kumar, Yalla, Natarajan,
  Chhaya, and Srinivasan}]{garimella_he_2021}
Aparna Garimella, Akhash Amarnath, K.~Kumar, Akash~Pramod Yalla, Anandhavelu
  Natarajan, Niyati Chhaya, and Balaji~Vasan Srinivasan. 2021.
\newblock \href {https://doi.org/10.18653/v1/2021.findings-acl.397} {He is very
  intelligent, she is very beautiful? {{On Mitigating Social Biases}} in
  {{Language Modelling}} and {{Generation}}}.
\newblock In \emph{{{FINDINGS}}}.

\bibitem[{Goyal et~al.(2022)Goyal, Li, and Durrett}]{Goyal2022NewsSA}
Tanya Goyal, Junyi~Jessy Li, and Greg Durrett. 2022.
\newblock News summarization and evaluation in the era of gpt-3.
\newblock \emph{ArXiv}, abs/2209.12356.

\bibitem[{Guo et~al.(2022)Guo, Ma, and Vosoughi}]{guo_measuring_2022}
Xiaobo Guo, Weicheng Ma, and Soroush Vosoughi. 2022.
\newblock Measuring {{Media Bias}} via {{Masked Language Modeling}}.
\newblock \emph{Proceedings of the International AAAI Conference on Web and
  Social Media}, 16:1404--1408.

\bibitem[{Hermann et~al.(2015)Hermann, Kocisky, Grefenstette, Espeholt, Kay,
  Suleyman, and Blunsom}]{hermann_teaching_2015}
Karl~Moritz Hermann, Tomas Kocisky, Edward Grefenstette, Lasse Espeholt, Will
  Kay, Mustafa Suleyman, and Phil Blunsom. 2015.
\newblock Teaching {{Machines}} to {{Read}} and {{Comprehend}}.
\newblock In \emph{Advances in {{Neural Information Processing Systems}}},
  volume~28. {Curran Associates, Inc.}

\bibitem[{Huang et~al.(2021)Huang, Feng, Feng, and Qin}]{Huang2021TheFI}
Yi-Chong Huang, Xiachong Feng, Xiaocheng Feng, and Bing Qin. 2021.
\newblock The factual inconsistency problem in abstractive text summarization:
  A survey.
\newblock \emph{ArXiv}, abs/2104.14839.

\bibitem[{Ji et~al.(2022)Ji, Lee, Frieske, Yu, Su, Xu, Ishii, Bang, Madotto,
  and Fung}]{ji2022survey}
Ziwei Ji, Nayeon Lee, Rita Frieske, Tiezheng Yu, Dan Su, Yan Xu, Etsuko Ishii,
  Yejin Bang, Andrea Madotto, and Pascale Fung. 2022.
\newblock Survey of hallucination in natural language generation.
\newblock \emph{arXiv preprint arXiv:2202.03629}.

\bibitem[{Kryscinski et~al.(2020)Kryscinski, McCann, Xiong, and
  Socher}]{kryscinski_evaluating_2020}
Wojciech Kryscinski, Bryan McCann, Caiming Xiong, and Richard Socher. 2020.
\newblock \href {https://doi.org/10.18653/v1/2020.emnlp-main.750} {Evaluating
  the {{Factual Consistency}} of {{Abstractive Text Summarization}}}.
\newblock In \emph{Proceedings of the 2020 {{Conference}} on {{Empirical
  Methods}} in {{Natural Language Processing}} ({{EMNLP}})}, pages 9332--9346,
  {Online}. {Association for Computational Linguistics}.

\bibitem[{Lee et~al.(2022)Lee, Bang, Yu, Madotto, and Fung}]{lee_neus_2022}
Nayeon Lee, Yejin Bang, Tiezheng Yu, Andrea Madotto, and Pascale Fung. 2022.
\newblock \href {https://doi.org/10.48550/arXiv.2204.04902} {{{NeuS}}:
  {{Neutral Multi-News Summarization}} for {{Mitigating Framing Bias}}}.
\newblock \emph{ArXiv}.

\bibitem[{Lewis et~al.(2019)Lewis, Liu, Goyal, Ghazvininejad, Mohamed, Levy,
  Stoyanov, and Zettlemoyer}]{lewis_bart_2019}
Mike Lewis, Yinhan Liu, Naman Goyal, Marjan Ghazvininejad, Abdelrahman Mohamed,
  Omer Levy, Ves Stoyanov, and Luke Zettlemoyer. 2019.
\newblock \href {http://arxiv.org/abs/1910.13461} {{{BART}}: {{Denoising
  Sequence-to-Sequence Pre-training}} for {{Natural Language Generation}},
  {{Translation}}, and {{Comprehension}}}.
\newblock \emph{arXiv:1910.13461 [cs, stat]}.

\bibitem[{Liang et~al.(2021)Liang, Wu, Morency, and
  Salakhutdinov}]{liang_towards_2021}
Paul~Pu Liang, Chiyu Wu, Louis-Philippe Morency, and R.~Salakhutdinov. 2021.
\newblock Towards {{Understanding}} and {{Mitigating Social Biases}} in
  {{Language Models}}.
\newblock In \emph{{{ICML}}}.

\bibitem[{Lin(2004)}]{lin_rouge_2004}
Chin-Yew Lin. 2004.
\newblock {{ROUGE}}: {{A Package}} for {{Automatic Evaluation}} of
  {{Summaries}}.
\newblock In \emph{Text {{Summarization Branches Out}}}, pages 74--81,
  {Barcelona, Spain}. {Association for Computational Linguistics}.

\bibitem[{Liu and Lapata(2019)}]{liu_text_2019}
Yang Liu and Mirella Lapata. 2019.
\newblock \href {http://arxiv.org/abs/1908.08345} {Text {{Summarization}} with
  {{Pretrained Encoders}}}.
\newblock \emph{arXiv:1908.08345 [cs]}.

\bibitem[{Maynez et~al.(2020{\natexlab{a}})Maynez, Narayan, Bohnet, and
  McDonald}]{maynez-etal-2020-faithfulness}
Joshua Maynez, Shashi Narayan, Bernd Bohnet, and Ryan McDonald.
  2020{\natexlab{a}}.
\newblock \href {https://doi.org/10.18653/v1/2020.acl-main.173} {On
  faithfulness and factuality in abstractive summarization}.
\newblock In \emph{Proceedings of the 58th Annual Meeting of the Association
  for Computational Linguistics}, pages 1906--1919, Online. Association for
  Computational Linguistics.

\bibitem[{Maynez et~al.(2020{\natexlab{b}})Maynez, Narayan, Bohnet, and
  McDonald}]{maynez_faithfulness_2020}
Joshua Maynez, Shashi Narayan, Bernd Bohnet, and Ryan McDonald.
  2020{\natexlab{b}}.
\newblock \href {https://doi.org/10.18653/v1/2020.acl-main.173} {On
  {{Faithfulness}} and {{Factuality}} in {{Abstractive Summarization}}}.
\newblock In \emph{Proceedings of the 58th {{Annual Meeting}} of the
  {{Association}} for {{Computational Linguistics}}}, pages 1906--1919,
  {Online}. {Association for Computational Linguistics}.

\bibitem[{Monroe et~al.(2009)Monroe, Colaresi, and Quinn}]{monroe_fightin_2009}
Burt Monroe, Michael Colaresi, and Kevin Quinn. 2009.
\newblock \href {https://doi.org/10.1093/pan/mpn018} {Fightin' {{Words}}:
  {{Lexical Feature Selection}} and {{Evaluation}} for {{Identifying}} the
  {{Content}} of {{Political Conflict}}}.
\newblock \emph{Political Analysis}, 16.

\bibitem[{Nan et~al.(2021)Nan, Nallapati, Wang, Nogueira~dos Santos, Zhu,
  Zhang, McKeown, and Xiang}]{nan-etal-2021-entity}
Feng Nan, Ramesh Nallapati, Zhiguo Wang, Cicero Nogueira~dos Santos, Henghui
  Zhu, Dejiao Zhang, Kathleen McKeown, and Bing Xiang. 2021.
\newblock \href {https://doi.org/10.18653/v1/2021.eacl-main.235} {Entity-level
  factual consistency of abstractive text summarization}.
\newblock In \emph{Proceedings of the 16th Conference of the European Chapter
  of the Association for Computational Linguistics: Main Volume}, pages
  2727--2733, Online. Association for Computational Linguistics.

\bibitem[{Nenkova and McKeown(2011)}]{nenkova2011automatic}
Ani Nenkova and Kathleen McKeown. 2011.
\newblock \emph{Automatic summarization}.
\newblock Now Publishers Inc.

\bibitem[{Niculae et~al.(2015)Niculae, Suen, Zhang, {Danescu-Niculescu-Mizil},
  and Leskovec}]{niculae_quotus_2015}
Vlad Niculae, Caroline Suen, Justine Zhang, Cristian {Danescu-Niculescu-Mizil},
  and J.~Leskovec. 2015.
\newblock \href {https://doi.org/10.1145/2736277.2741688} {{{QUOTUS}}: {{The
  Structure}} of {{Political Media Coverage}} as {{Revealed}} by {{Quoting
  Patterns}}}.
\newblock \emph{WWW}.

\bibitem[{Patel and Pavlick(2021)}]{patel-pavlick-2021-stated}
Roma Patel and Ellie Pavlick. 2021.
\newblock \href {https://doi.org/10.18653/v1/2021.emnlp-main.790} {{``}was it
  {``}stated{''} or was it {``}claimed{''}?: How linguistic bias affects
  generative language models}.
\newblock In \emph{Proceedings of the 2021 Conference on Empirical Methods in
  Natural Language Processing}, pages 10080--10095, Online and Punta Cana,
  Dominican Republic. Association for Computational Linguistics.

\bibitem[{Qi et~al.(2020)Qi, Yan, Gong, Liu, Duan, Chen, Zhang, and
  Zhou}]{qi_prophetnet_2020}
Weizhen Qi, Yu~Yan, Yeyun Gong, Dayiheng Liu, Nan Duan, Jiusheng Chen, Ruofei
  Zhang, and Ming Zhou. 2020.
\newblock \href {http://arxiv.org/abs/2001.04063} {{{ProphetNet}}: {{Predicting
  Future N-gram}} for {{Sequence-to-Sequence Pre-training}}}.
\newblock \emph{arXiv:2001.04063 [cs]}.

\bibitem[{Ratcliff and Metzener(1988)}]{ratcliff1988pattern}
John~W Ratcliff and David~E Metzener. 1988.
\newblock Pattern {Matching}: {The} {Gestalt} {Approach}.
\newblock \emph{Dr Dobbs Journal}, 13(7):46.

\bibitem[{Shandilya et~al.(2018)Shandilya, Ghosh, and
  Ghosh}]{shandilya_fairness_2018}
Anurag Shandilya, Kripabandhu Ghosh, and Saptarshi Ghosh. 2018.
\newblock \href {https://doi.org/10.1145/3184558.3186947} {Fairness of
  {{Extractive Text Summarization}}}.
\newblock In \emph{Companion {{Proceedings}} of the {{The Web Conference}}
  2018}, {{WWW}} '18, pages 97--98, {Republic and Canton of Geneva, CHE}.
  {International World Wide Web Conferences Steering Committee}.

\bibitem[{Sheng et~al.(2019)Sheng, Chang, Natarajan, and
  Peng}]{sheng_woman_2019}
Emily Sheng, Kai-Wei Chang, P.~Natarajan, and Nanyun Peng. 2019.
\newblock \href {https://doi.org/10.18653/v1/D19-1339} {The {{Woman Worked}} as
  a {{Babysitter}}: {{On Biases}} in {{Language Generation}}}.
\newblock \emph{EMNLP}.

\bibitem[{Sheng et~al.(2021)Sheng, Chang, Natarajan, and
  Peng}]{sheng-etal-2021-societal}
Emily Sheng, Kai-Wei Chang, Prem Natarajan, and Nanyun Peng. 2021.
\newblock \href {https://doi.org/10.18653/v1/2021.acl-long.330} {Societal
  biases in language generation: Progress and challenges}.
\newblock In \emph{Proceedings of ACL}, pages 4275--4293, Online. Association
  for Computational Linguistics.

\bibitem[{Steen and Markert(2023)}]{steen2023investigating}
Julius Steen and Katja Markert. 2023.
\newblock \href {http://arxiv.org/abs/2309.08047} {Investigating gender bias in
  news summarization}.

\bibitem[{Sunstein and Lessig(1994)}]{sunstein_president_1994}
Cass Sunstein and Lawrence Lessig. 1994.
\newblock The {{President}} and the {{Administration}}.
\newblock \emph{Columbia Law Review}, 94:1.

\bibitem[{Tan et~al.(2018)Tan, Peng, and Smith}]{tan+peng+smith:18}
Chenhao Tan, Hao Peng, and Noah~A. Smith. 2018.
\newblock ``you are no jack kennedy'': On media selection of highlights from
  presidential debates.
\newblock In \emph{Proceedings of WWW}.

\bibitem[{Tang et~al.(2022)Tang, Goyal, Fabbri, Laban, Xu, Yahvuz,
  Kry{\'s}ci{\'n}ski, Rousseau, and Durrett}]{tang_understanding_2022}
Liyan Tang, Tanya Goyal, Alexander~R. Fabbri, Philippe Laban, Jiacheng Xu,
  Semih Yahvuz, Wojciech Kry{\'s}ci{\'n}ski, Justin~F. Rousseau, and Greg
  Durrett. 2022.
\newblock \href {https://doi.org/10.48550/arXiv.2205.12854} {Understanding
  {{Factual Errors}} in {{Summarization}}: {{Errors}}, {{Summarizers}},
  {{Datasets}}, {{Error Detectors}}}.

\bibitem[{Vallejo et~al.(2023)Vallejo, Baldwin, and
  Frermann}]{vallejo2023connecting}
Gisela Vallejo, Timothy Baldwin, and Lea Frermann. 2023.
\newblock \href {http://arxiv.org/abs/2309.08069} {Connecting the dots in news
  analysis: A cross-disciplinary survey of media bias and framing}.

\bibitem[{Vig et~al.(2020)Vig, Gehrmann, Belinkov, Qian, Nevo, Sakenis, Huang,
  Singer, and Shieber}]{vig_causal_2020}
Jesse Vig, Sebastian Gehrmann, Yonatan Belinkov, Sharon Qian, Daniel Nevo,
  Simas Sakenis, Jason Huang, Yaron Singer, and Stuart Shieber. 2020.
\newblock \href {https://doi.org/10.48550/arXiv.2004.12265} {Causal {{Mediation
  Analysis}} for {{Interpreting Neural NLP}}: {{The Case}} of {{Gender Bias}}}.

\bibitem[{Wang et~al.(2022)Wang, Rubinstein, and Cohn}]{wang_measuring_2022}
Jun Wang, Benjamin Rubinstein, and Trevor Cohn. 2022.
\newblock \href {https://doi.org/10.18653/v1/2022.acl-long.184} {Measuring and
  {{Mitigating Name Biases}} in {{Neural Machine Translation}}}.
\newblock In \emph{Proceedings of the 60th {{Annual Meeting}} of the
  {{Association}} for {{Computational Linguistics}} ({{Volume}} 1: {{Long
  Papers}})}, pages 2576--2590, {Dublin, Ireland}. {Association for
  Computational Linguistics}.

\bibitem[{Wilson(2016)}]{wilson2016barack}
John~K Wilson. 2016.
\newblock \emph{Barack Obama: This Improbable Quest}.
\newblock Routledge.

\bibitem[{Yang et~al.(2023)Yang, Li, Zhang, Chen, and
  Cheng}]{Yang2023ExploringTL}
Xianjun Yang, Yan Li, Xinlu Zhang, Haifeng Chen, and Wei Cheng. 2023.
\newblock Exploring the limits of chatgpt for query or aspect-based text
  summarization.
\newblock \emph{ArXiv}, abs/2302.08081.

\bibitem[{Yeo and Chen(2020)}]{yeo_defining_2020}
C.~Yeo and A.~Chen. 2020.
\newblock \href {https://doi.org/10.18653/v1/2020.winlp-1.27} {Defining and
  {{Evaluating Fair Natural Language Generation}}}.
\newblock \emph{WINLP}.

\bibitem[{Zhang et~al.(2023)Zhang, Liu, and Zhang}]{Zhang2023ExtractiveSV}
Haopeng Zhang, Xiao Liu, and Jiawei Zhang. 2023.
\newblock Extractive summarization via chatgpt for faithful summary generation.
\newblock \emph{ArXiv}, abs/2304.04193.

\bibitem[{Zhang et~al.(2020)Zhang, Zhao, Saleh, and Liu}]{zhang_pegasus_2020}
Jingqing Zhang, Yao Zhao, Mohammad Saleh, and Peter~J. Liu. 2020.
\newblock \href {http://arxiv.org/abs/1912.08777} {{{PEGASUS}}:
  {{Pre-training}} with {{Extracted Gap-sentences}} for {{Abstractive
  Summarization}}}.
\newblock \emph{arXiv:1912.08777 [cs]}.

\end{thebibliography}
